\documentclass[preprint,12pt]{elsarticle}
\usepackage{graphicx}
\usepackage{amssymb}

\newtheorem{theo}{\bf Theorem}%[section]
\newtheorem{defi}{\bf Definition}%[section]
\newtheorem{lemm}{\bf Lemma}%[section]
\begin{document}
\begin{frontmatter}
\title{Removing Mixture of Gaussian and Impulse Noise by Patch-Based Weighted Means}

 \author[1,2]{Haijuan Hu} 
\ead{huhaijuan61@126.com }
\author[3]{Bing Li } 
\ead{libcs@263.net}
\author[2]{Quansheng Liu}
\ead{Quansheng.Liu@univ-ubs.fr}

\address[1]{Northeastern University at Qinhuangdao,
              School of Mathematics and Statistics,  Hebei, 066004, China}
\address[2]{Univ Bretagne-Sud, CNRS UMR 6205,  LMBA, Campus de Tohannic, F-56000 Vannes, France
}
\address[3]{Zhongshan Polytechnic, Department of Math., Zhongshan, 528404, China}

%%\author{ Haijuan Hu Bing Li and Quansheng Liu},Universit\'e Europ\'eenne de Bretagne, 
%%\address

\begin{abstract}
 We first establish a law of large numbers and a convergence theorem in distribution to show the rate of convergence of  the non-local means filter for removing Gaussian noise. We then introduce the notion of degree of similarity to measure the role of similarity for the non-local means filter.  Based on the convergence theorems,  we  propose a patch-based weighted means filter for removing impulse noise and its  mixture with Gaussian noise by combining the essential idea of the trilateral filter and that of the non-local means filter. Our experiments show that our filter is competitive compared to recently proposed methods. 

\end{abstract}


\end{frontmatter}

%\maketitle

\section{Introduction}
%$\mathcal{N}  \mathbb{R}$
Images are produced to record or display useful information. Due to the visibility of images and the rapid development of  science and technology, images play an increasingly important role in our lives.
% processing has an explosive influence on our modern society. 
However,  because of imperfections in the imaging and capturing process,  the recorded image invariably represents a degraded 
version of the original scene (\cite{bovik2005handbook,kenneth1996castleman}). The undoing of these imperfections 
is crucial to many of the subsequent image processing tasks. 

There exists a wide range of different degradations. A very important example   is the existence of noise.  Noise may be introduced by the medium
through which the image is created and transmitted.  In this paper, we concentrate on removing impulse noise and its mixture with Gaussian noise.

We present a numerical image  by a $M\times N$ matrix $u = \{u(i):
i\in I \}$, where $I=\{0, 1,\dots, M-1\} \times \{0, 1, \dots, N-1\}$ is the image domain,
and $ u(i) \in \{0,1,2,\dots, 255\}$ represents the gray value at the pixel $i$ for 8-bit gray images.  The additive Gaussian noise
model is:
$$v(i)=u(i)+\eta(i), $$
 where $u=\{ u(i): i\in I\}$ is the
original image, $v=\{ v(i): i\in I\}$ is the noisy one, and $\eta$ is the
Gaussian noise: $\eta(i)$ are independent and identically distributed
Gaussian random variables with mean $0$ and standard deviation
 $\sigma$. In the sequel we always denote by $u$ the original image, and $v$ the
noisy one.  The random  impulse noise model is:

 $$v(i)= \left\{\begin{array}{ll}
\eta(i) & \mbox{with probability} \,\, p, \\
u(i) & \mbox{with probability} \,\, (1-p), \end{array} \right.     $$
where $0<p<1$ is the
 impulse probability (the proportion of the occurrence of the impulse noise),
and $\eta(i)$ are independent random variables uniformly distributed on  $[\mbox{min}\{u(i):i\in I\},\mbox{max}\{u(i):i\in I\}]$, generally taken as [0,255] for 8-bit gray images. The task of image denoising is to recover the unknown original image $u$ as well as possible from the degraded one $v$.

 There are a lot of denoising methods in the literature. 
To remove  Gaussian noise, there are approaches based on wavelets \cite{donoho1994ideal,coifman1995translation,chang2000adaptive}, approaches based on variational models \cite{rudin_nonlinear_1992,durand2003reconstruction}, and weighted means approaches \cite{yaroslavsky1985digital,smith1997susan,tomasi1998bilateral,buades2005review,kervrann2008local}, etc.  A very important progress  is the proposition of the non-local means filter (NL-means) \cite{buades2005review}, which estimates  original images by weighted means along similar local patches.  Since then,  many researchers combined the ideas of ``local patch" and other methods to remove noise, for instance \cite{elad2006image,dabov2007image,mairal2008learning,hu2012nonlocal}. 
 There are also  many methods to remove impulse noise including, for example,  median based filters \cite{pratt1975median,chen2001adaptive,akkoul2010new}, fuzzy filters \cite{yuksel2006hybrid}, variational based methods \cite{nikolova_variational_2004,chan2004iterative,dong2007detection}. 

Because  Gaussian noise and impulse noise have different characters, the above-mentioned methods can only  be used for removing only one kind of noise (Gaussian noise or impulse noise), and  can not be applied to remove a mixture of Gaussian noise and impulse noise. To remove mixed noise, a successful method is the trilateral filter  \cite{garnett2005universal}: it introduces  an interesting statistic called
ROAD (Rank of Ordered Absolute Differences)  to detect impulse noisy pixels, and  combines it with the bilateral filter \cite{smith1997susan,tomasi1998bilateral} to remove noise.  The trilateral filter \cite{garnett2005universal} is also effective for removing impulse noise; a variant of the ROAD statistic,  named ROLD (Rank-Ordered Logarithmic Difference),  has been proposed in  \cite{dong2007detection}, where it is  combined with the edge-preserving variational method \cite{chan2004iterative} for removing impulse noise.  
Other methods have also been developed  recently  to remove mixed noise. The papers \cite{li2011new,huliliu,xiong2012universal,delon2012patch,delon2013patch}  use the patch-based idea of NL-means to remove impulse noise and mixed noise.
In \cite{li2011new,huliliu},    generalizations of NL-means are proposed for removing impulse noise and its mixture with Gaussian noise  with the help of the  ROAD statistics \cite{garnett2005universal}; the main idea is to define weights in terms of the ROAD statistics and the similarity of local patches, which are nearly zero for impulse noisy points. 
 In \cite{xiong2012universal},  NL-means is adapted   by estimating the similarity of patches   with the reference image obtained in an impulse noise detection mechanism.  The papers  \cite{delon2012patch,delon2013patch} also use a patch-based approach: they 
introduce a robust distance inspired by order statistics to estimate the similarity between patches  using the tail of the binomial distribution,  and use the maximum likelihood estimator (MLE) to estimate  the original image.
 The methods in \cite{yang2009mixed} and \cite{xiao2011restoration,zhou2013restoration}  use the ideas of the state-of-the-art Gaussian noise removal  methods BM3D \cite{dabov2007image} and K-SVD \cite{elad2006image} respectively to remove mixed noise; the algorithm proposed in \cite{lopez2010restoration} is based on a
Bayesian classification of the input pixels, which is combined with the kernel regression framework.

The success of NL-means \cite{buades2005review} explores in a nice way the similarity phenomenon which is widespread in natural images. 
Many filters have been developed based on the basic idea of NL-means as stated before. But the theoretic aspects have not been so much studied. A probabilistic explanation called similarity principle is given in \cite{li2011new}. Here we  improve this principle by proving a Marcinkiewicz law of large numbers and a convergence theorem in distribution, which describe the rate of convergence of NL-means.  We will also introduce the notion of degree of similarity to estimate the influence of similarity on NL-means. Based on the convergence theorems, we propose a new filter, called patch-based weighted means filter (PWMF) to improve the mixed noise filter (MNF) introduced in \cite{li2011new}. Compared to MNF, the new filter  simplifies the joint impulse factor in MNF, adds a spatial factor, and adjusts the choice of parameters.  Experimental results show that our new filter is  competitive both for removing impulse noise and mixed noise compared to recently developed filters \cite{garnett2005universal,yang2009mixed,xiao2011restoration,xiong2012universal,delon2013patch}. For the sake of  completeness, let us mention that this paper is an extended and  improved version of the conference proceeding version  \cite{huliliu}.

This paper  is organized as follows. In Section 2, we recall the  non-local means filter \cite{buades2005review},  establish two  convergence theorems to show the rate of convergence  of NL-means, and introduce the notion of degree of similarity. In Section 3, we recall the trilateral filter \cite{garnett2005universal} and introduce our new filter. Experiments are presented in Section 4. Conclusions are made in Section 5.

\section{{Convergence Theorems for  Non-Local Means   }}\label{mainths}
In this section, we first establish a Marcinkiewicz law of large numbers and a convergence  theorem in distribution for  NL-means \cite{buades2005review}. We then introduce the notion of degree of similarity to measure the role of similarity in  image restoration. 
\subsection{Non-Local Means}
 For $i \in I$ and $d$  an odd integer, let
$\mathcal{N}_{i}(d)= \{j\in I: |j-i| \leq (d-1)/2\}$  be the window with center
$i$ and size $d \times d$, where  $|j-i| = \max(|j_1-i_1|, |j_2-i_2|)$ for $i=(i_1,i_2)$ and $j=(j_1,j_2)$.
And set
$\mathcal{N}_{i}^{0}(d)=\mathcal{N}_{i}(d)\backslash\{i\}$. We sometimes simply write
$\mathcal{N}_{i}$ and $\mathcal{N}_{i}^{0}$ for
 $\mathcal{N}_{i}(d)$ and $\mathcal{N}_{i}^{0}(d)$, respectively. Denote  $v(\mathcal{N}_{i})=\{v(k):k \in \mathcal{N}_{i}\}$ as the vector composed of the gray values of $v$ in the window $\mathcal{N}_{i}$ arranged lexicographically.

The denoised image by NL-means is given by
\begin{equation}
\mbox{NLM}(v) (i) =\frac{\sum_{j\in \mathcal{N}_i(D)} w(i,j)v(j)}{\sum_{j\in \mathcal{N}_i(D)} w(i,j)}, \quad i\in I,
\end{equation}
with
\begin{equation}
 w(i,j) = e^{- ||v(\mathcal{N}_{i})-v(\mathcal{N}_{j} )||_a^{2}/ (2\sigma_r^2)}, \label{nlmdis}
\end{equation} 
where  $\sigma_r>0$ is a control parameter,
\begin{equation}
     ||v(\mathcal{N}_{i})-v(\mathcal{N}_{j})||_a^2 =\frac{ \sum_{k\in \mathcal{N}_i(d)} a(i,k) |v(k)-v(\mathcal{T}(k))|^2}{\sum_{k\in \mathcal{N}_i(d)} a(i,k)}, \label{nlmdisnorm}
\end{equation}
 $ a(i,k)>0$ being some fixed weights usually chosen  to be
a decreasing function of the Euclidean norm $\|i-k\|$ or $|i-k|$, and  $\mathcal{T}=\mathcal{T}_{ij}$ is the translation mapping of $\mathcal{N}_i$ onto $\mathcal{N}_j$:  $\mathcal{T}(k)=k-i+j, k\in \mathcal{N}_i.$ 
Originally, $\mathcal{N}_{i}(D)$ in $(\ref{nlmdis})$ is chosen as the whole image $I$, but in practice, it is better to choose $\mathcal{N}_{i}(D)$ with an appropriate number $D$. We call $\mathcal{N}_{i}(D)$ search windows, and $\mathcal{N}_i=\mathcal{N}_{i}(d)$ local patches. 

\subsection{Convergence Theorems }
\subsubsection{Convergence Theorems for NL-means}\label{mainth}
We now present two convergence theorems for NL-means using probability theory. 
For simplicity, we use the same notation $v(\mathcal{N}_{i})$ to denote both the observed image patch centered at $i$  and the  corresponding random variable (in fact the observed image is just a realization of the corresponding variable). Therefore the distribution of the observed image $v(\mathcal{N}_{i})$ is just that of the corresponding random variable.

\begin{defi}   Two patches $v(\mathcal{N}_{i})$  and $v(\mathcal{N}_{j})$   are called similar if they have the same probability distribution.    
\label{defi1}
\end{defi}

Definition \ref{defi1} is a probabilistic interpretation of the similarity phenomenon. According to this definition, two observed patches  $v(\mathcal{N}_{i})$  and $v(\mathcal{N}_{j})$ are similar if they are issued from the same probability distribution. 

The following theorem is a kind of  Marcinkiewicz law of  large numbers. It gives an estimation of the almost sure convergence rate of the estimator to the original image for the non-local means filter.

\begin{theo}
Let $i\in I $ and 
let $I_i$ be the set of $j$ such that the patches $v(\mathcal{N}_{i})$  and $v(\mathcal{N}_{j})$ are similar (in the sense of Definition $\ref{defi1}$). Set 
\begin{equation}
 {v^{0}(i)}=\frac{\sum_{j\in I_i}{w^{0}}(i,j)v(j)}{\sum_{j\in I_i}{w^{0}}(i,j)},\label{v0i}
\end{equation}
where
\begin{equation}
{w^{0}(i,j)}=e^{-\|v({\mathcal{N}_i^0})-v(\mathcal{N}_j^0)\|_a^2/(2\sigma_r^2)}.
\label{wei0}
\end{equation}
Then for any  $\epsilon\in(0,\frac12]$, as $|I_i|\rightarrow\infty$,
\begin{equation} 
 v^0(i)-u(i)=o(|I_i|^{-(\frac12-\epsilon)})  \quad\mbox{almost surely}, \label{th21}
\end{equation}\label{th1}
where $|I_i|$ denotes the cardinality of $I_i$.
\end{theo}

Notice that when $\epsilon=\frac12$, ($\ref{th21}$) means that 
\begin{equation} 
 \lim_{|I_i|\to\infty}v^0(i)=u(i)  \quad\mbox{almost surely},\label{th21sim}
\end{equation}
which is the similarity principle in \cite{li2011new}. 

Recall that  $\mathcal{N}_i^0=\mathcal{N}_i\backslash{i}$.
Theorem $\ref{th1}$ shows that $v^0(i)$ is a good estimator of the original image $u(i)$ if the number of similar patches   $|I_i|$ is sufficiently large. Here we use the weight $w^0 (i,j)$ instead of $w(i,j)$,  as 
  $w^0 (i,j)$ has the nice property that it is independent of $ v(j)$ if $j\not\in \mathcal{N}_{i}$. This property is used in the proof, and makes the estimator $v^0 (i)$ to be ``almost'' non-biased: in fact, if the family $\{v(j)\}_j$ is independent of the family  $\{w^0(i,j)\}_j $ (e.g. this is the case when the similar patches are disjoint), then it is evident that $\mathbb{E}v^0(i) = u(i)$. We can consider that this non-biased property holds approximately as for each $j$ there are few $k$ such that $w^0 (i,k)$ is dependent of $v(j)$. A closely related  explanation  about  the biased  estimation of  NL-means  can be found in \cite{xu2008biased}.  

 Notice that when $v(\mathcal{N}_{j})$ is not similar to $v(\mathcal{N}_{i})$,  the weight $w^0(i,j)$   is small and  negligible. Therefore it is also reasonable to take all patches for the calculation. But selecting only similar patches can improve the restoration result, and can also speed up the algorithm.
The difference between $ w^0(i,j)$  and $w(i,j)$ is also small, so that Theorem $\ref{th1}$  shows that NLM$(v)(i)$ is also a good estimator of $u(i)$. But very often   $v^0(i)$ gives better restoration result.

The next result is a convergence theorem in distribution. It states  that $v^0(i)-u(i) \to 0$ in distribution with a rate as $1/\sqrt{|I_i|}$.

\begin{theo} Under the  condition of Theorem $\ref{th1}$, assume additionally  that 
 $\{v(\mathcal{N}_{j}): j\in I_i\}$  is a stationary sequence of random vectors. 
Then as $  |I_i|\to\infty$,
$$\sqrt{|I_i|}\big(v^0(i)-u(i)\big)\stackrel{d}{\to} \mathcal{L}, $$  %\xi,{I_i} 
where $\stackrel{d}{\to}$ means the convergence in distribution, 
and $\mathcal{L}$ is a mixture of centered Gaussian laws in the sense that it has a density of the form
$$ f(t)=\int_{\mathbb{R}^{|\mathcal{N}_i^0|}}\frac1{\sqrt{2\pi}c_x}e^{-\frac{t^2}{2c_x^2}}\mu(dx),$$
$\mu$ being the law of $v(\mathcal{N}_i^0)$ and $c_x>0$. 
\label{th2}
\end{theo}

By Theorems $\ref{th1}$ and $\ref{th2}$, the larger the value of $|I_i|$, the better the approximation of $v^0(i)$ to $u(i)$, that is to say, the more similar patches, the better restored result.
 This will be confirmed  
in Section \ref{secsim} where we shall introduce the notion of degree of   similarity   for images, showing that the larger the degree of similarity, the better the quality of restoration. 
 The proofs of the theorems will be given in Section \ref{proofthm}.

To prove the two theorems,  we will show a Marcinkiewicz law of  large numbers (Theorem $\ref{genth1}$) and a convergence theorem in distribution for random weighted means (Theorem $\ref{genth2}$),  
 which are more general than Theorems $\ref{th1}$ and  $\ref{th2}$, respectively.
\subsubsection{Convergence Theorems for Random Weighted Means}
 Before stating the theorems, we first recall the notion of $l$-dependence. %give a definition. 
\begin{defi} For an integer $l\geq0$, a sequence of random variables $X_1, X_2, \dots$ is called to be $l$-dependent if each subsequence $X_{k_1}, X_{k_2}, \dots$ is independent whenever $|k_m-k_n|>l$ for all $m,n\geq 1.$ (That is, random variables with distances greater than $l$ are  independent of each other.) 
\end{defi}

\begin{theo}
Let $\{(a_k, v_k)\}$ be a sequence of $l$-dependent identically distributed random variables, with $\mathbb{E}|a_1|^p<\infty$ and $ \mathbb{E}|a_1v_1|^p<\infty$ for some $ p\in[1,2),$ and $\mathbb{E} a_1\neq 0$. Then
$$ \frac{\sum_{k=1}^na_kv_k}{\sum_{k=1}^na_k}-\frac{\mathbb{E}a_1v_1}{\mathbb{E}a_1}=o(n^{-(1-1/p)}) \quad\mbox{almost surely}.$$
\label{genth1}
\end{theo}

We need the following lemma to prove it.

\begin{lemm}\cite{li2011new} 
If $\{X_n\}$ are $l$-dependent and identically distributed random variables with $ \mathbb{E}X_1=0$ and $\mathbb{E}|X_1|^p<\infty$ for some $p\in [1,2)$, then
$$\lim_{n\rightarrow\infty}\frac{X_1+\dots+X_n}{n^{1/p} }=0 \quad \mbox{almost surely}.$$
\label{lemm1}
\end{lemm}

This lemma is a direct consequence of Marcinkiewicz law of large numbers for independent random
 variables \cite{chow2003probability}(Page 118), since for all $k\in\{1,\dots,l+1\}, \{X_{i(l+1)+k}:i\geq 0\}$
 is a sequence of i.i.d. random variables, and for each positive integer $n$, we have
$$ X_1+\dots+X_n=\sum_{k=1}^{l+1}\sum_{i=0}^{m-1}X_{i(l+1)+k}+\sum_{1\leq k\leq k_0}X_{m(l+1)+k},
$$
where $m,k_0$ are positive integers with $n=m(l+1)+k_0, 0\leq k \leq l$.

\emph{Proof of Theorem $\ref{genth1}$.}
Notice that $$n \bigg(\frac{\sum_{k=1}^na_kv_k}{\sum_{k=1}^na_k}-\frac{\mathbb{E}a_1v_1}{\mathbb{E}a_1}\bigg)= \frac{n}{(\sum_{k=1}^na_k)}  \frac{1}{\mathbb{E}a_1} \sum_{k=1}^n a_k z_k,$$
where $$ z_k=v_k \mathbb{E}a_1-\mathbb{E}a_1 v_1.$$
Since $\mathbb{E}|a_1|\leq (\mathbb{E}|a_1|^p)^{1/p}<\infty$, by Lemma $\ref{lemm1}$ with $p=1$, we have
$$\lim_{n\rightarrow\infty}\frac{\sum_{k=1}^na_k}n=\mathbb{E}a_1 \quad\mbox{almost surely}.$$
Since $\mathbb{E}a_1z_1=0$, and $\mathbb{E}|a_1z_1|^p<\infty$, again by Lemma 
$\ref{lemm1}$, we get
$$\lim_{n\rightarrow\infty}\frac{\sum_{k=1}^na_kz_k}{n^{1/p}}=0 \quad\mbox{almost surely}.$$
Thus the conclusion follows.
\hfill $\Box$

\begin{theo} Let $\{(a_k,v_k)\}$ be a stationary sequence of $l$-dependent and identically distributed random variables with $\mathbb{E}a_1\neq0, \mathbb{E}a_1^2<\infty,$ and $\mathbb{E}(a_1v_1)^2<\infty$. Then
$$ {\sqrt n} \bigg(\frac{\sum_{k=1}^na_kv_k}{\sum_{k=1}^na_k}-\frac{\mathbb{E}a_1v_1}{\mathbb{E}a_1}\bigg)\stackrel{d}{\to} N(0,c^2),$$
 that is,
$$\lim_{n\to \infty}\mathbb{P}\bigg\{\frac{\sqrt n}{c} \bigg(\frac{\sum_{k=1}^na_kv_k}{\sum_{k=1}^na_k}-\frac{\mathbb{E}a_1v_1}{\mathbb{E}a_1}\bigg)\leq z\bigg\}=\Phi(z), \quad z\in \mathbb{R}, $$
where
$$ \Phi(z)=\frac1{\sqrt{2\pi}}\int_{-\infty}^z
e^{-\frac{t^2}2}dt,$$  
and $c$ is a constant.
\label{genth2}
\end{theo}

We need the following lemma to prove the theorem.

\begin{lemm}\cite{wen1986sur}
 Let $\{X_n\}$ be a stationary sequence of $l$-dependent and identically distributed random variables with  $\mathbb{E}X_1=0$ and  $\mathbb{E}X_1^2=c^2<\infty $.  Set $S_n=X_1+\dots+X_n (n\geq1)$,  %%then we have var$(S_n)=An+B,$ where
$$ c_1=c^2+2\sum_{k=1}^l\mathbb{E}X_1X_{1+k}, \quad\mbox{and} \quad c_2=2\sum_{k=1}^lk\mathbb{E}X_1X_{1+k}.$$ 
 Then \mbox{var}$(S_n)=c_1n-c_2$, 
and as $n\to\infty$,$$ \frac{S_n}{\sqrt{\mbox{var}(S_n)}}\stackrel{d}{\to}N(0,1).$$
\label{lemm2}
\end{lemm}

\emph {Proof of Theorem $\ref{genth2}$.  }
As in the proof of Theorem $\ref{genth1}$, we have
$${\sqrt{n}} \bigg(\frac{\sum_{k=1}^na_kv_k}{\sum_{k=1}^na_k}-\frac{\mathbb{E}a_1v_1}{\mathbb{E}a_1}\bigg)=
\frac{n}{(\sum_{k=1}^na_k)}  \frac{1}{\mathbb{E}a_1} \frac{\sum_{k=1}^n a_k z_k}{\sqrt{n}}, $$
where $$ z_k=v_k \mathbb{E}a_1-\mathbb{E}a_1 v_1.$$
Notice that the $l$-dependence of $\{(a_k,v_k)\}$ and the stationarity  imply those of $\{(a_k,z_k)\}$. Therefore by  Lemma $\ref{lemm2}$, we get 
$$  \frac{\sum_{k=1}^n a_k z_k}{\sqrt{n}c_0}\to N(0,1),$$
where $$c_0=\sqrt{\mathbb{E}(a_1z_1)^2+2\lambda}, \quad \mbox{with}\, \lambda=\sum_{k=1}^l\mathbb{E}a_1z_1a_{1+k}z_{1+k}.$$
Since $\mathbb{E}|a_1|\leq (\mathbb{E}|a_1|^p)^{1/p}$, by Lemma $\ref{lemm1}$ with $p=1$, we obtain $$\lim_{n\rightarrow\infty}\frac{\sum_{k=1}^na_k}n=\mathbb{E}a_1 \quad\mbox{almost surely}.$$
Thus the conclusion follows with $c=c_0 (\mathbb{E}a_1)^2$. \hfill $\Box$

\subsubsection{Proofs of the Convergence  Theorems $\ref{th1}$ and $\ref{th2}$ } \label{proofthm}

For simplicity, denote by $I_{i}=\{j_1,j_2,\dots\, j_n\}$ with $n=|I_i|$ in the following proof. We need to prove, for any  $\epsilon\in(0,\frac12]$, as $n\rightarrow\infty$,
$$\frac{\sum_{k=1}^n{w^{0}}(i,j_k)v(j_k)}{\sum_{k=1}^n{w^{0}}(i,j_k)}-u(i)=o(n^{-(\frac12-\epsilon)})  \quad\mbox{almost surely},
$$
with
$$w^{0}(i,j_k)=e^{-\|v(\mathcal{N}^{0}_{i})-v(\mathcal{N}^{0}_{j_k})\|^2/(2\sigma_r^2)}.
$$
We will apply Theorem $\ref{genth1}$ to prove it.  Note that the sequence $\{{w^{0}}(i,j_k),v(j_k)\}$ ${(k=1,2,\dots,n)}$ is not $l$-dependent, since the central random variable $v(\mathcal{N}^{0}_{i})$ is contained in all the terms. To make use of Theorem $\ref{genth1}$, we first take a fixed vector to replace the central random variable.

\emph {Proof of Theorem $\ref{th1}$.}
 Fix $x\in \mathbb{R}^{|\mathcal{N}_i^0|}$.  Let $$a_{k}=w^0(x,j_k)=e^{-\|x-v(\mathcal{N}^{0}_{j_k})\|^2/(2\sigma_r^2)},$$ then $a_{k}$ and $v(j_k)$ are independent since $j_k\not\in \mathcal{N}^{0}_{j_k}$. So $$\frac {\mathbb{E}a_{k}v(j_k)}{\mathbb{E}a_{k}}=\mathbb{E}v(j_k)=u(j_k)=u(i).$$
When $\mathcal{N}^{0}_{j_m}$ and $\mathcal{N}^{0}_{j_n}$ are disjoint, $v(\mathcal{N}^{0}_{j_m})$ and $v(\mathcal{N}^{0}_{j_n})$ are independent, so are $a_{j_m}$ and $a_{j_n}$, and so are $\big(a_{j_m},v(j_m)\big)$ and $\big(a_{j_n},v(j_n)\big)$ since $x$ is fixed. Thus $\{\big(a_{k},v(j_k)\big)\}$ is a sequence of $l$-dependent identically distributed random vectors for $l$ large enough.  
Since $v=u+\eta$, with the range of $u$ being bounded  and $\eta$ being Gaussian, we have $\mathbb{E}|v(j_k)|^p<\infty$ for $p\in [1,2)$. (In fact, it holds for all $p\geq 1$.) Thus $\mathbb{E}|a_{k}v(j_k)|^p<\infty$, as $a_k\leq1$. 

Applying  Theorem $\ref{genth1}$, for fixed $x=v(\mathcal{N}^0_i)\in \mathbb{R}^{|\mathcal{N}_i^0|}$,  we have, for any positive integer   $k_0$,

\begin{equation} \frac{\sum_{k=k_0}^n{w^{0}}(x,j_k)v(j_k)}{\sum_{k=k_0}^n{w^{0}}(x,j_k)}-u(i)=o(n^{-(1-1/p})\quad\mbox{almost surely}.  \label{pf211}   \end{equation}
Now let $k_0$ be a positive integer  such  that $v(\mathcal{N}_i^0)$ is independent of $v(\mathcal{N}^{0}_{j_k})$  for all  $k\geq k_0$. 
By  Fubini's theorem, we can replace $w^{0}(x,j_k)$ in $(\ref{pf211})$ by $$w^{0}(i,j_k)=e^{-\|v(\mathcal{N}^{0}_{i})-v(\mathcal{N}^{0}_{j_k})\|^2/(2\sigma_r^2)}.$$ That is,

\begin{equation} \frac{\sum_{k=k_0}^n{w^{0}}(i,j_k)v(j_k)}{\sum_{k=k_0}^n{w^{0}}(i,j_k)}-u(i)=o(n^{-(1-1/p)})\quad\mbox{almost surely}.  \label{pf212} \end{equation}

To prove the theorem, we need to estimate the difference between the left-hand sides of  $(\ref{th21})$ and  $(\ref{pf212})$.
Let $$
A_0=\sum_{k=1}^{k_0-1}{w^{0}}(i,j_k)v(j_k), \quad A_n=\sum_{k=k_0}^n{w^{0}}(i,j_k)v(j_k),
$$
$$
B_0=\sum_{k=1}^{k_0-1}{w^{0}}(i,j_k), \quad B_n=\sum_{k=k_0}^n{w^{0}}(i,j_k).
$$
Then as before,  fixing $x\in \mathbb{R}^{|\mathcal{N}_i^0|}$, applying Theorem $\ref{genth1}$ with $p=1$ and Fubini's theorem, and replacing $x$ by $v(\mathcal{N}_i^0)$, we obtain
$$
\lim_{n\to\infty} \frac{A_n}n=\mathbb{E}{w^{0}}(i,j_k)v(j_k), \quad \mbox{and} \lim_{n	\to\infty} \frac{B_n}n=\mathbb{E}{w^{0}}(i,j_k).
$$
Using this and the fact that%(\ref{th21})-(\ref{pf212})=|(\ref{th21})-(\ref{pf212})|=
$$
\frac{A_0+A_n}{B_0+B_n}-\frac{A_n}{B_n}=\frac{A_0B_n-A_nB_0}{B_n(B_0+B_n)},
$$
we  see that
\begin{equation}
\bigg|\frac{\sum_{k=k_0}^n{w^{0}}(i,j_k)v(j_k)}{\sum_{k=k_0}^n{w^{0}}(i,j_k)}-\frac{\sum_{k=1}^n{w^{0}}(i,j_k)v(j_k)}{\sum_{k=1}^n{w^{0}}(i,j_k)}\bigg|=O(\frac1n) \quad \mbox{almost surely}. \label{diffe}
\end{equation}
Therefore, ($\ref{pf212}$) implies that 
\begin{equation} \frac{\sum_{k=1}^n{w^{0}}(i,j_k)v(j_k)}{\sum_{k=1}^n{w^{0}}(i,j_k)}-u(i)=o(n^{-(1-1/p)})\quad\mbox{almost surely}. \label{pff}  \end{equation}
As ($\ref{pff}$) holds for any $p \in [1,2)$, we see that ($\ref{th21}$) holds for all $\epsilon\in(0,\frac12]$.
\hfill $\Box$

\vskip 5mm

\emph {Proof of Theorem $\ref{th2}$.} 
Similarly to the proof of Theorem $
\ref{th1}$, fix $x\in \mathbb{R}^{|\mathcal{N}_i^0|}$, and set $a_{k}=w^0(x,j_k)=e^{-\|x-v(\mathcal{N}^{0}_{j_k})\|^2/(2\sigma_r^2)}$. Then $a_{k}$ and $v(j_k)$ are independent, and $ {\mathbb{E}a_{k}v(j_k)}/{\mathbb{E}a_{k}}=\mathbb{E}v(j_k)=u(i)$. For $l$ large enough, $\{(a_{k},v(j_k))\}$ is a sequence of $l$-dependent identically distributed random vectors,  and $ \mathbb{E}|a_{k}v(j_k)|^2\leq \mathbb{E}|v(j_k)|^2<\infty$. %Denote $\eta=\{(v(\mathcal{N}_{j_k}): k\geq k_0\}$. 
 Hence applying  Theorem $\ref{genth2}$, for fixed $x=v(\mathcal{N}^0_i)$,
we get, for any positive integer   $k_0$,
$$ Z_n(x):={\sqrt n} \bigg(\frac{\sum_{k=k_0}^n{w^{0}}(x,j_k)v(j_k)}{\sum_{k=k_0}^n{w^{0}}(x,j_k)}-u(i)\bigg)\stackrel{d}{\to} N(0,c_x^2),$$
where $c_x>0$ depends on $x$.
This means that for any $t\in\mathbb{R}$,
$$
\lim_{n\to\infty}\mathbb{P}(Z_n(x) \leq t)=\int_{-\infty}^t\frac1{\sqrt{2\pi}c_x}e^{-\frac{z^2}{2c_x^2}}dz.
$$
Let $k_0$ be the positive integer  such  that $v(\mathcal{N}_i^0)$ is independent of  $v(\mathcal{N}^{0}_{j_k})$  for all $k\geq k_0$. 
Then 
by Fubini's theorem and Lebesgue's dominated convergence theorem, we have
$$
\lim_{n\to\infty}\mathbb{P}\Big(Z_n\big(v(\mathcal{N}^0_i)\big) \leq t\Big)=\int_{-\infty}^t f(z) dz,
$$
where $$ f(z)=\int_{\mathbb{R}^{|\mathcal{N}_i^0|}}\frac1{\sqrt{2\pi}c_x}e^{-\frac{z^2}{2c_x^2}}\mu(dx),$$
with $\mu$ being the law of $v(\mathcal{N}_i^0)$. In other words, 
$$ {\sqrt n} \bigg(\frac{\sum_{k=k_0}^n{w^{0}}(i,j_k)v(j_k)}{\sum_{k=k_0}^n{w^{0}}(i,j_k)}-u(i)\bigg)\stackrel{d}{\to} \mathcal{L},$$
where $\mathcal{L}$ is the law with density $f$. This together with ($\ref{diffe}$) give the conclusion of Theorem $
\ref{th2}$.

\hfill $\Box$

\subsection{{Degree of Similarity}} \label{secsim}
\noindent In this section, we introduce the notion of degree of similarity of images to describe the influence of the number of similar patches on the restoration results.  

 If  two patches $v(\mathcal{N}_{i})$ and $v(\mathcal{N}_j)$ %are similar (that is, if they 
have  the same distribution and  are independent of each other, then $\{v(k)-v(\mathcal{T}(k)): k\in\mathcal{N}_{i}\}$ are independent variables with the same
law $N(0,2\sigma^2)$ (recall that $\mathcal{T}$ is the translation mapping of $\mathcal{N}_i$ onto $\mathcal{N}_j$).  Therefore
$$\|v(\mathcal{N}_{i})-v(\mathcal{N}_j)\|^2=\sum_{k\in \mathcal{N}_{i}}|v(k)-v(\mathcal{T}(k))|^2=2\sigma^2X,$$
where $X$ is the sum of squares of independent random variables with normal law $N(0,1)$, and  has the law $\chi^2$ with $|\mathcal{N}_{i}|=d^2$
degrees of liberty. %_{\alpha}
For $\alpha\in(0,1)$, let $T_{\alpha}>0$ be determined by 
$$
\mathbb{P}(X>\frac{T_{\alpha}^2}{2\sigma^2})=\alpha,
$$
so that $$ \mathbb{P}(\|v(\mathcal{N}_{i})-v(\mathcal{N}_j)\|\leq T_{\alpha})=1-\alpha.$$
The value of $\alpha$ is chosen to be small enough. It represents the risk probability. When $\mathbb{P}(\|v(\mathcal{N}_{i})-v(\mathcal{N}_j)\|\leq T_{\alpha}$, we consider that $v(\mathcal{N}_{i})$ and $v(\mathcal{N}_{j})$ are similar with confidence level $1-\alpha$. This leads us to the following definition of the degree of similarity.

\begin{defi}
For $i\in I,$ let
$$ DS_i=\frac{\#\{j\in \mathcal{N}_{i}(D):\|v(\mathcal{N}_{i})-v(\mathcal{N}_j)\|\leq T_{\alpha}\}}{D^2}$$
be the proportion of  patches  $v(\mathcal{N}_j)$ similar to $v(\mathcal{N}_{i})$ in the search window $\mathcal{N}_{i}(D)$, and let
$$ DS=\frac{\sum_{i\in I}DS_i}{MN}$$ be their mean over the whole image. Define $DS$ the degree of
similarity of the image $v$ with confidence level $1-\alpha$.
\end{defi}

We compute the DS values for different  images corrupted by Gaussian noise with $\sigma=10, 20, 30$. Note that the DS values depend on $\alpha, d, D, \sigma$. In our experiments, we fix $\alpha=0.1$. The DS values of noisy images are shown in Table $\ref{tabsim}$.  It can be seen that, for non-local means based methods, for example, our method in Section \ref{pwmf}, generally, the larger the DS value, 
the larger the PSNR value (see Section \ref{exp}).

\begin{table}[h]
\begin{center}%Table I.\;\;
\caption{ DS values of noisy images ($\alpha=0.1, d=9$) } 
 {\footnotesize
 \begin{tabular}{ccc|c|c|c|c|c}
\hline
$\sigma$ & $T_{\alpha}$  &$D$ & Lena & Bridge &Peppers256  & Peppers512 & Boats \\
\hline
$10$ & $111.39$  &$7$ & 0.42 & 0.06 & 0.29 & 0.39 & 0.25 \\ 
\hline
$20$ &$279.54$  &$11$ & 0.54 & 0.15 & 0.39 & 0.54 & 0.40 \\
\hline
$30$& $589.33$  &$15$ & 0.55 & 0.18 & 0.38 & 0.55 & 0.42 \\

\end{tabular}
}

\label{tabsim}
\end {center}
\end{table}

\section{Patch-Based Weighted Means Filter }
In this section, we  introduce our new filter which combines the basic idea of NL-means \cite{buades2005review} and that of the trilateral filter \cite{garnett2005universal}.   

\subsection{Trilateral Filter}
The authors of  \cite{garnett2005universal} proposed a neighborhood filter called the trilateral filter as an extension of the bilateral filter \cite{smith1997susan,tomasi1998bilateral}  to remove random impulse noise and its mixture with 
Gaussian noise. Firstly, they introduced the statistic ROAD (Rank of Ordered Absolute Differences) to measure how like a point is an impulse noisy point defined by 
\begin{equation}
\mbox {ROAD} (i) = r_{1}(i)+\cdots+r_{m}(i), \label{road}
\end{equation}
$r_{k}(i)$ being the $k$-th smallest term
 in the set $\{ | u(i)-u(j)|: j\in \mathcal{N}_{i}(d)\backslash\{i\}\}$,
  $d$ and $m$ two constants taken as $d=1, m=4$ in \cite{garnett2005universal}. 
If $i$  is an impulse noisy point, then ROAD$(i)$ is large; otherwise it is small. Therefore, the  ROAD statistic serves to  detect impulse noisy points.

Secondly, with the ROAD statistic, they defined
 the impulse  factor $w_{I}(i)$ and the
joint impulse factor $J_I(i,j)$:
\begin{eqnarray}%{rl}
 w_{I}(i) & =&e^{-\frac{\mbox{\tiny ROAD}(i)^{2}}{2\sigma_{I}^{2}}},  \label{wii} \\
 J_I(i,j) & =&e^{-\Big(\frac{\big(\mbox{\tiny ROAD}(i)+\mbox{\tiny ROAD}(j)\big)}2\Big)^{2}\big/(2\sigma_{J}^{2})}, \label{ji}
\end{eqnarray}
where $\sigma_{I}$ and $\sigma_{J}$ are control parameters\footnote{In fact, \cite{garnett2005universal} defines the joint impulse factor as $J(i,j)=1-J_I(i,j)$.  However, it seems to be more convenient to use $J_I(i,j)$ than $J(i,j)$ \cite{li2011new}.}. 
If $i$ is an impulse noisy point, then the value of $w_{I}(i)$ is close to 0; otherwise it is close to 1. Similarly, 
if either $i$ or $j$ is   an impulse noisy point, then the value of $J_I(i,j)$ is close to $0$; otherwise it is close to 1.

Finally, the restored image by the trilateral filter is
%\begin{equation*}
\begin{equation}
 \mbox {TriF}(v)(i) =\frac{\sum_{j\in \mathcal{N}_{i}(D)} w(i,j)v(j)}{\sum_{j\in \mathcal{N}_{i}(D)} w(i,j)},
 \end{equation}
where

%\begin{equation*}\end{equation*}\begin{equation*}\end{equation*}\begin{array}{ll}
$$
w(i,j)=w_{S}(i,j)w_R(i,j)^{J_I(i,j)}w_I(j)^{1-J_I(i,j)},$$
with $$ w_{S}(i,j)=e^{-\frac{|i-j|^2}{2\sigma_{S}^2}},\quad
w_{R}(i,j) =e^{-\frac{(v(i)-v(j))^2}{2\sigma_{R}^2}}.
$$

\subsection{Patch-Based Weighted Means Filter} \label{pwmf}
As in the non-local means filter \cite{buades2005review}, our filter estimates each point by the weighed means of its neighbors, and the weight for each neighbor is determined by the similarity of local patches centered at the estimated point and the neighbor. Due to the existence of impulse noise, some points are totally destroyed, so that noisy values are not related to  original values at all. So we have to diminish the influence of impulse noisy points. 
Similarly to \cite{li2011new,huliliu}, we introduce the following  weighted norm: 

\begin{equation}
%\begin{array}{ll}
 ||v(\mathcal{N}_{i})-v(\mathcal{N}_{j} )||_{M}^{2} =
\frac
 {\displaystyle{\sum_{k\in \mathcal{N}_{i}^0} w_{S,M}(i,k) F\big(k,\mathcal{T}(k)\big)\, |v(k)-v\big(\mathcal{T}(k)\big)|^{2}}}
 {\displaystyle{\sum_{k\in \mathcal{N}_{i}^0} w_{S,M}(i,k)F\big(k,\mathcal{T}(k)\big)}},\label{impnorm}
%\end{array}
\end{equation}
where \begin{equation}
w_{S,M}(i,k)=e^{-\frac{|i-k|^2}{2\sigma_{S,M}^2}}, \quad F\big(k,\mathcal{T}(k)\big)=w_I(k)w_I\big(\mathcal{T}(k)\big). 
\label{sigsm}
\end{equation}
Recall that here $k=(k_1, k_2)$ represents a two-dimensional spatial location of a pixel, $w_{I}$ is defined in ($\ref{wii}$), and $\mathcal{T}$ is the translation mapping of $\mathcal{N}_i(d)$ onto $\mathcal{N}_j(d)$. 
$F\big(k,\mathcal{T}(k)\big)$ is a joint impulse factor: if $k$ or $\mathcal{T}(k)$ is an impulse noisy point, then  $F\big(k,\mathcal{T}(k)\big)$ is close to $0$,
so that these points contribute little to the weighted norm; otherwise $F\big(k,\mathcal{T}(k)\big)$ is close to $1$. 

We now define our  filter that we call {\em Patch-based Weighted Means
Filter} (PWMF). The restored image by PWMF is defined
as
\begin{equation}
 \mbox {PWMF}(v)(i) =\frac{\sum_{j\in \mathcal{N}_{i}(D)} w(i,j)v(j)}{\sum_{j\in \mathcal{N}_{i}(D)} w(i,j)} , 
 \end{equation}
  where
$$
  w(i,j)=w_{S}(i,j)w_{I}(j)w_{M}(i,j),$$
with \begin{equation}
 w_{S}(i,j)=e^{-{|i-j|^2}/{(2\sigma_{S}^2})},\quad
  w_{M}(i,j)=e^{-{||v(\mathcal{N}_{i})-v(\mathcal{N}_{j})||_{M}^{2}}/(2\sigma_{M}^{2})}, \label{sigssigm}
\end{equation}
and $w_{I}(j)$ is defined in ($\ref{wii}$). 
By definition, for each impulse noisy point $j$ in  $\mathcal{N}_{i}(D)$,  $w(i,j)$ is close to $0$. Hence our
new filter can be regarded as an application of the mathematical justifications of the non-local means filter stated in Section \ref{mainths} 
to the remained image (which can be considered to contain only Gaussian noise) obtained after filtering the impulse
noisy points by the   weighted norm (\ref{impnorm}).

Finally, we mention that, in this paper, we use the joint impulse factor $F\big(k,\mathcal{T}(k)\big)$ $=w_I(k)w_I\big(\mathcal{T}(k)\big)$, which is different from the choice in \cite{li2011new} and \cite{huliliu}, where $F\big(k,\mathcal{T}(k)\big)$ $=J_I\big(k,\mathcal{T}(k)\big)$. In fact, we can see that  with this new choice, we simplify the methods in \cite{li2011new} and \cite{huliliu} by eliminating a parameter and speeding up the implementation. Furthermore, we empirically find that the new choice leads to an improvement of the quality of restored images, especially for impulse noise.

\section{Simulations}

\subsection{Choices of Parameters}

Notice that PWMF  reduces to NL-means when $\sigma_I= \sigma_S=\infty$. So for removing Gaussian noise, a reasonable choice is to take  $\sigma_I$ and  $\sigma_S$ large enough.
Now we present the choices of parameters for removing impulse noise and mixed noise, which are important for our filter.  

In the calculation of ROAD (cf.(\ref{road})), we choose $3\times3$ neighborhoods and $m=4$.  For impulse noise or mixed noise with $p=0.4, 0.5$, to further improve the results,   $5\times 5$ neighborhoods and $m=12$ are used to calculate  ROAD. Now we come to  the choice of $\sigma_I$, $\sigma_M$, $\sigma_{S}$, $\sigma_{S,M}$ appearing in (\ref{wii}),  (\ref{sigssigm}) and (\ref{sigsm}). To remove impulse noise, we use $\sigma_M=3+20p$, $\sigma_{S}=0.6+p$, and omit the factor  $w_{S,M}$ (i.e. $\sigma_{S,M}$ can be taken a value large enough);  $\sigma_I=50$ for $p=0.2, 0.3$, and $\sigma_I=160$ for $p=0.4, 0.5$. For mixed noise, we use $\sigma_I=50+5\sigma/3$, $\sigma_M=3+0.4\sigma+20p$, $\sigma_{S,M}=2$, and omit the factor $w_{S}$. The patch size $d=9$ is used in all cases; the  search window sizes  $D$ are shown in Table \ref{nlmixfpara}. For other values of $\sigma$ or $p$, we choose  parameters by linear interpolation or  according to the adjacent values of $\sigma$ or $p$.  It is not  easy to find appropriate parameters for a filter. Different choices of parameters can have great influence to the restored images.    See also \cite{delon2013patch}.  Some similar research for NL-means can be found in \cite{xu2008biased,van2009sure,duval2011bias}. Note that our choice of parameters is different from   \cite{li2011new}: for the patch size, we use $d=9$, while   \cite{li2011new} uses $d=3$ in most cases; for impulse noise with $p=0.4,0.5$, we use $5\times 5$  neighborhoods for ROAD, while \cite{li2011new} always uses $3\times 3$  neighborhoods.

\begin{table}
\begin{center}
\caption{Choice of search window sizes $D$ for PWMF}
 {\footnotesize
 \begin{tabular}{c|c|c|c|c}
\hline
  & $ \sigma= 0 $ &$ \sigma=10 $ & $ \sigma= 20 $ & $ \sigma=  30$ \\
\hline
$D$& $7$ & $ 7 $ & $ 11 $ & $ 15 $ \\
\end{tabular}}
\label{nlmixfpara}
\end{center}
\end{table}

\subsection{Experiments and Comparisons} \label{exp}
%image source, cf. \hfill \\

\begin{figure}
\begin{center}

\includegraphics[width=0.45\linewidth]{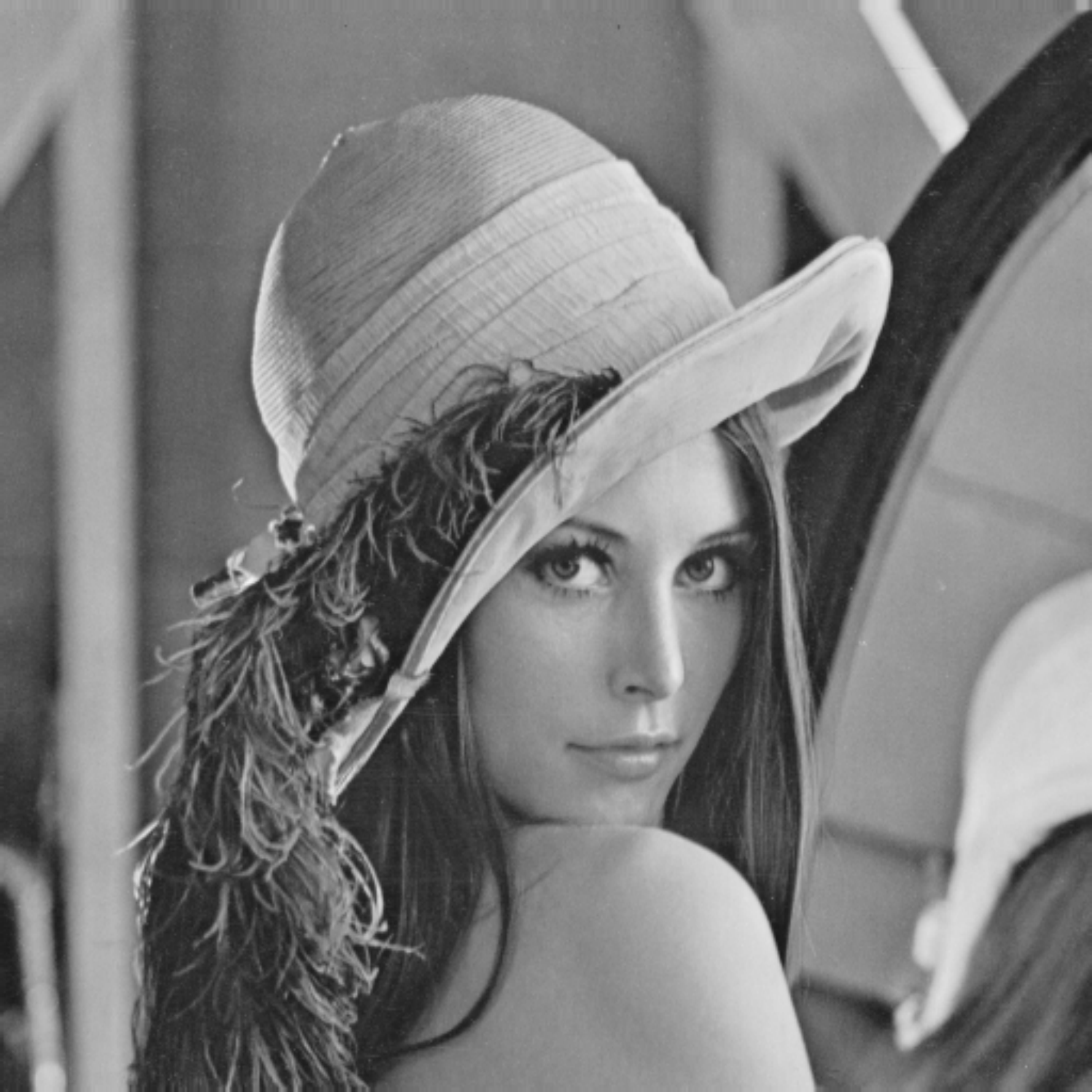}  
\includegraphics[width=0.45\linewidth]{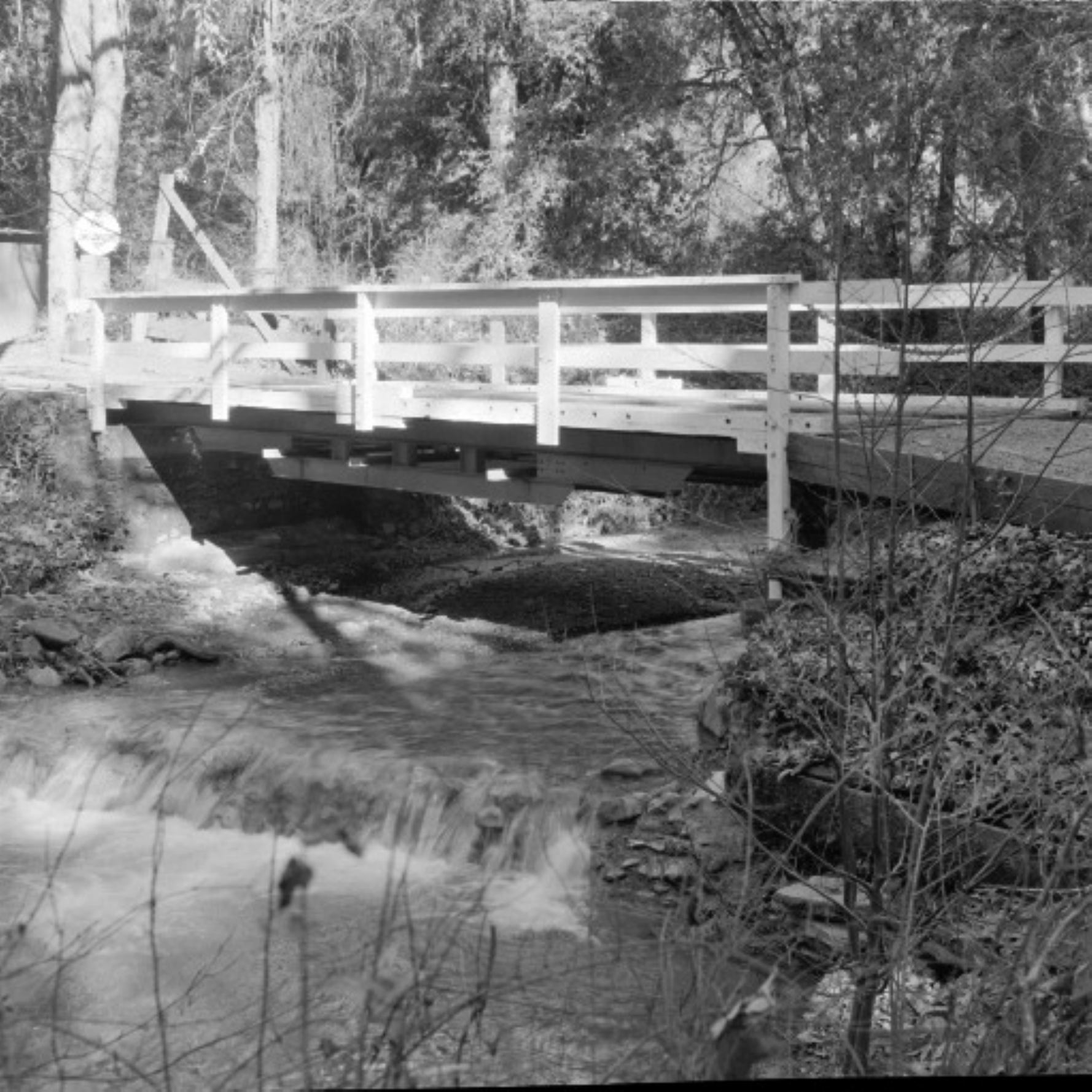}  
\includegraphics[width=0.45\linewidth]{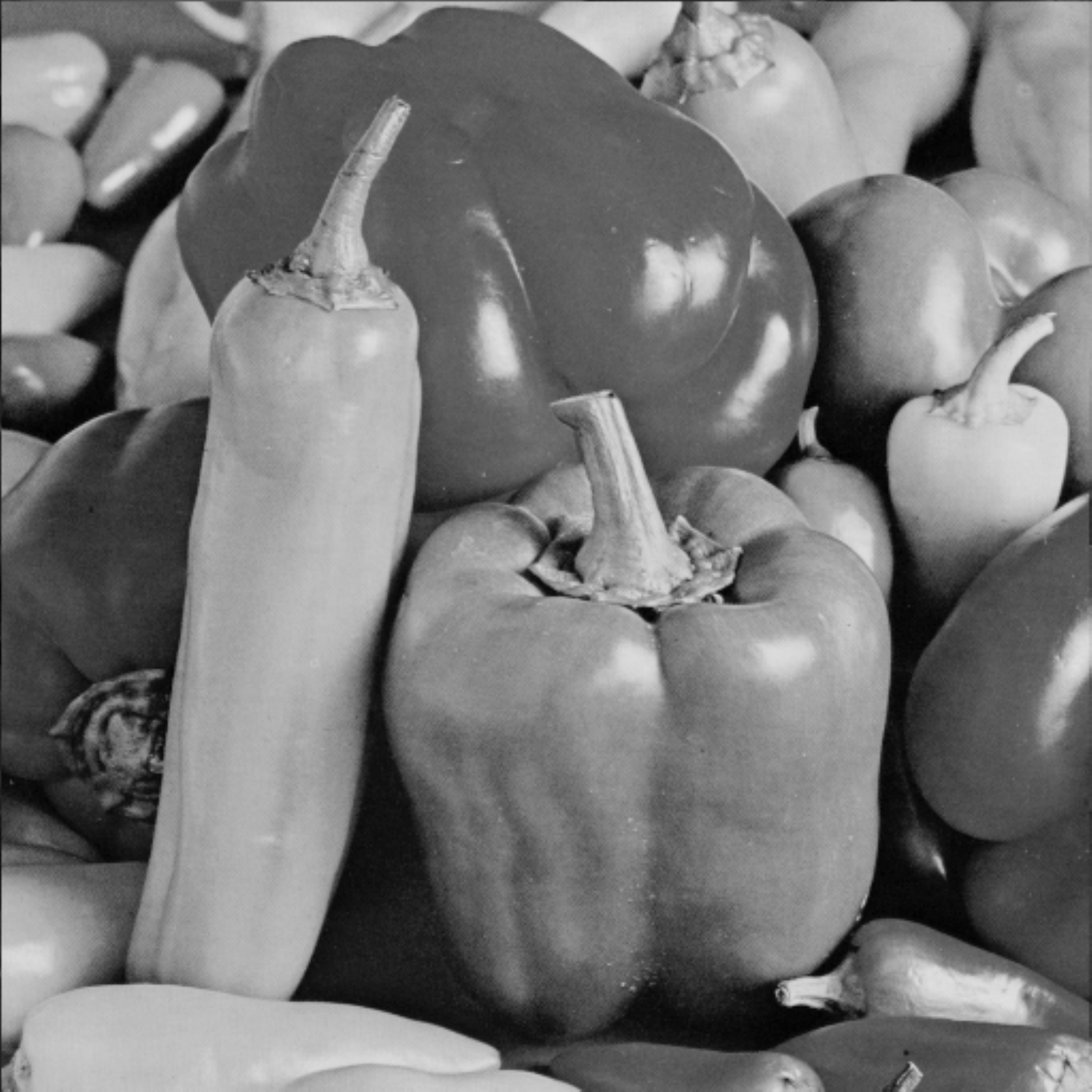}
\includegraphics[width=0.45\linewidth]{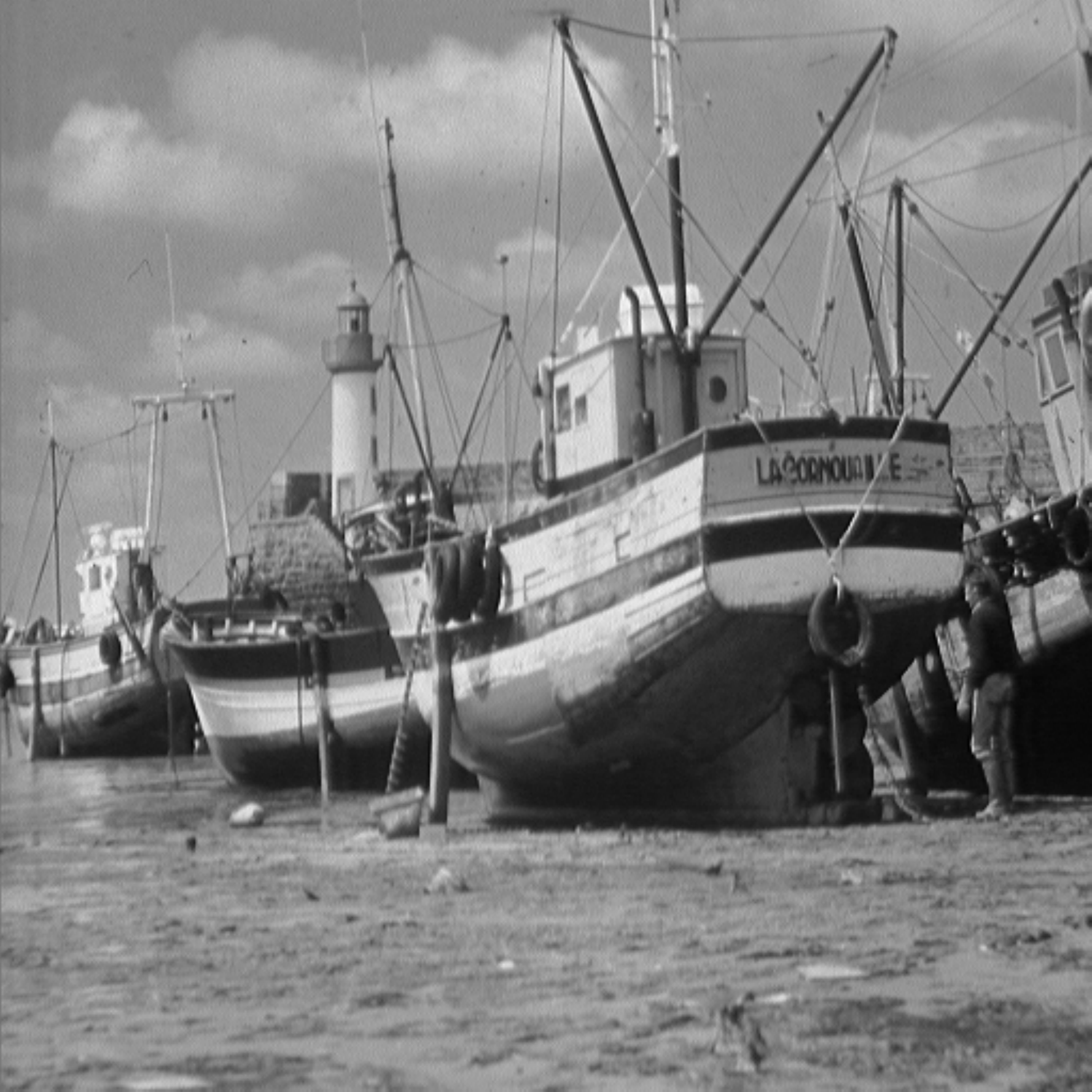}  

\caption{ Original $512\times 512$ images of Lena, Bridge, Peppers512, Boats. Since in the original  Peppers images, there are  black boundaries of width of one pixel in the left  and top  which can be considered as impulse noise, to make an impartial comparison, we compute PSNR for Peppers images  after removing all the four boundaries, that is with images of size $510\times 510$ for Peppers512 and $254\times254$ for Peppers256}
\label{original}%and patch size $d$   
\end{center}
\end{figure}

We use  standard gray images to test the performance of our filter\footnote{The code of our method and the images can be downloaded at \hfill \\ https://www.dropbox.com/s/oylg9to8n6029hh/to\_j\_sci\_comput\_paper\_code.zip.}. Original images are shown in Fig. \ref{original}.\footnote{
The images Lena, Peppers256 and Boats are originally downloaded  from \hfill \\ http://decsai.ugr.es/$\sim$javier/denoise/test\_images/index.htm;

the image Peppers512 is from http://perso.telecom-paristech.fr/$\sim$delon/Demos/Impulse

 and the image Bridge is from  www.math.cuhk.edu.hk/$\sim$rchan/paper/dcx/.} 
As usual we use PSNR (Peak Signal-to-Noise Ratio)
$$
  \mbox{PSNR } (\bar {v}) = 10\log_{10} \frac{255^2|I|}{\sum_{i\in I }(\bar v(i) - u(i))^2} \mbox{dB}
  $$ 
   to measure the quality of a restored image, 
  where  $u$ is the   original image,  and $\bar v$ the restored one. For the simulations, the gray value of impulse noise is uniformly distributed on the interval [0,255]. We add  Gaussian noise  and then add impulse noise for
the simulation of mixed noise. We  use same realizations of noisy images for comparisons of different methods when  codes are available, that is, for TriF \cite{garnett2005universal}, ROLD-EPR \cite{dong2007detection}, NLMixF \cite{huliliu}  and MNF \cite{li2011new}. For other methods, we list the reported results in papers.

The results for TriF are obtained by the  program made by ourselves. 
 To compare the performance of our filter with those of TriF fairly, we  make our effort to obtain the best results as we can according to the suggestion  of \cite{garnett2005universal}.  
 We use  $\sigma_I=40, \sigma_J=50, \sigma_S=0.5,$ and $\sigma_R=2\sigma_{QGN}$, where $\sigma_{QGN}$ is an estimator for the standard deviation of ``quasi-Gaussian'' noise defined in \cite{garnett2005universal}.
For impulse noise, 
  we apply one iteration for $p=0.2$, two iterations for $p=0.3, 0.4$, and  four iterations for $p=0.5$.
For mixed noise, we   apply TriF twice with different
 values of $\sigma_S$ as suggested in \cite{garnett2005universal}: with all impulse noise levels $p$, for $\sigma=10$, we  first use $\sigma_{S}=0.3$, then $\sigma_{S}=1$; for $\sigma=20$, first $\sigma_{S}=0.3$, then $\sigma_{S}=15$; for $\sigma=30$,  first $\sigma_{S}=15$, then $\sigma_{S}=15$. %Note that when $\sigma_{S}=15$,   we can omit the spatial factor.

For ROLD-EPR, the listed values are the best PSNR values along iterations with the code from the authors of \cite{dong2007detection}.

\begin{figure}
\begin{center}
\begin{tabular}{ccc} %left bottom right up
\includegraphics[trim = 5cm 5cm 5cm 5cm, clip, width=0.30\linewidth]{fig11_21.pdf}&
\includegraphics[trim = 5cm 5cm 5cm 5cm, clip, width=0.30\linewidth]{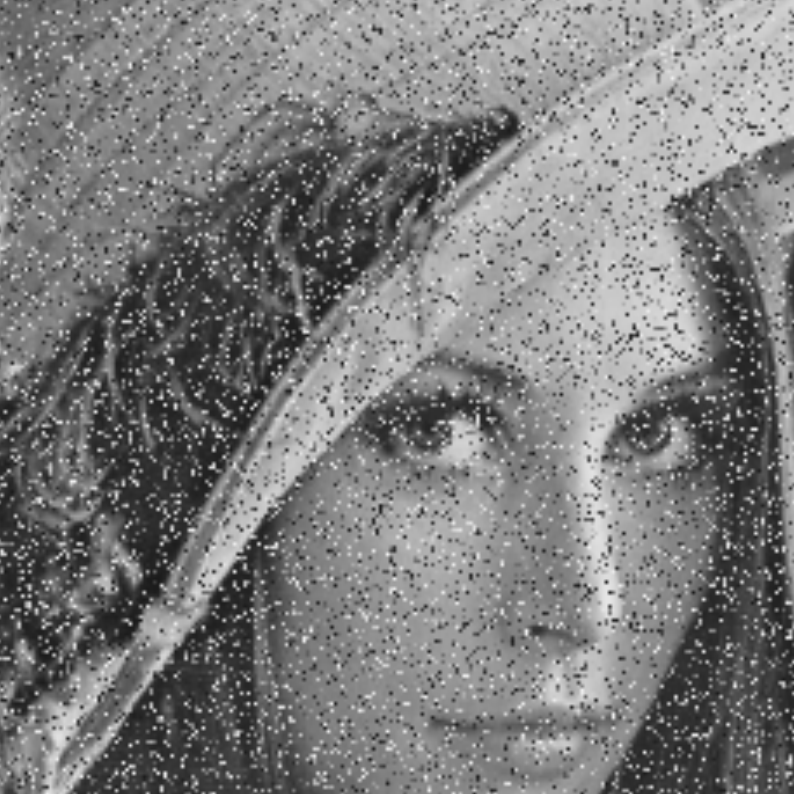} 
&\includegraphics[trim = 5cm 5cm 5cm 5cm, clip, width=0.30\linewidth]{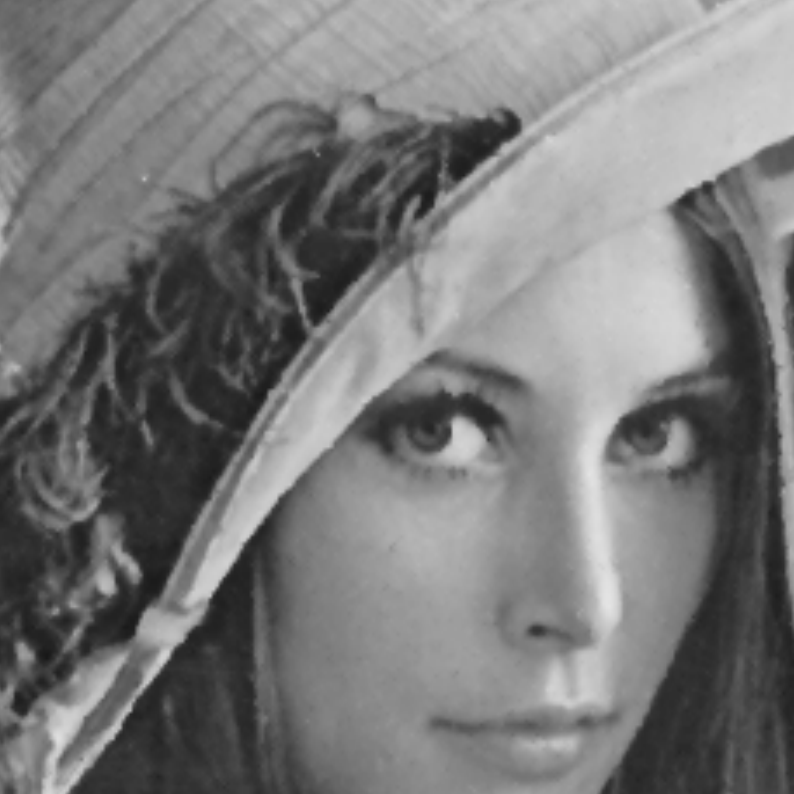}
\\

{\small Original } & {\small Noisy $p$ = 0.2 }& {\small TriF PSNR = 34.75 }\\
\includegraphics[trim = 5cm 5cm 5cm 5cm, clip, width=0.30\linewidth]{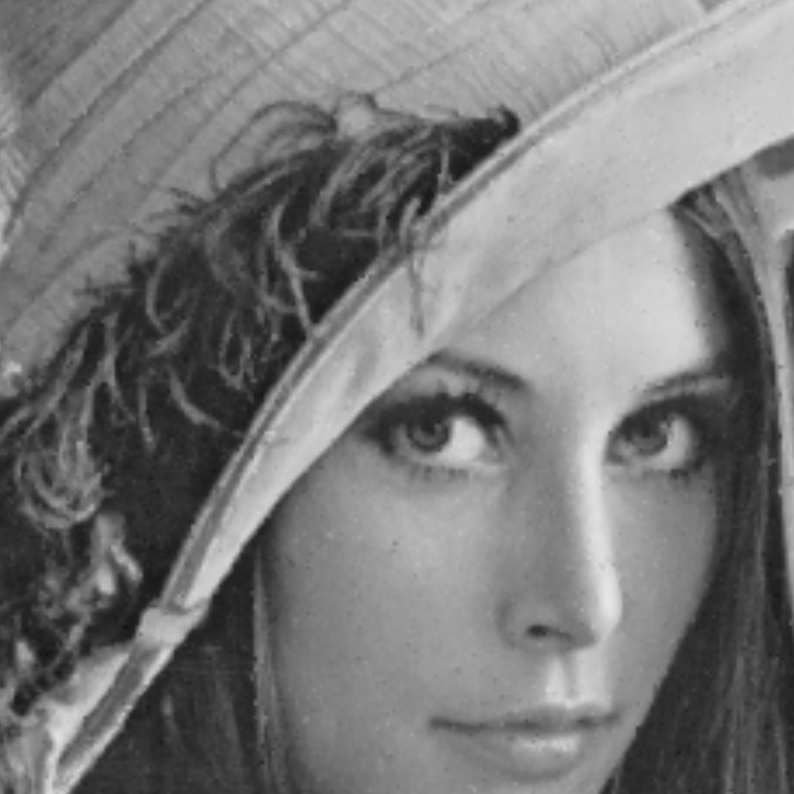}
&\includegraphics[trim = 5cm 5cm 5cm 5cm, clip, width=0.30\linewidth]{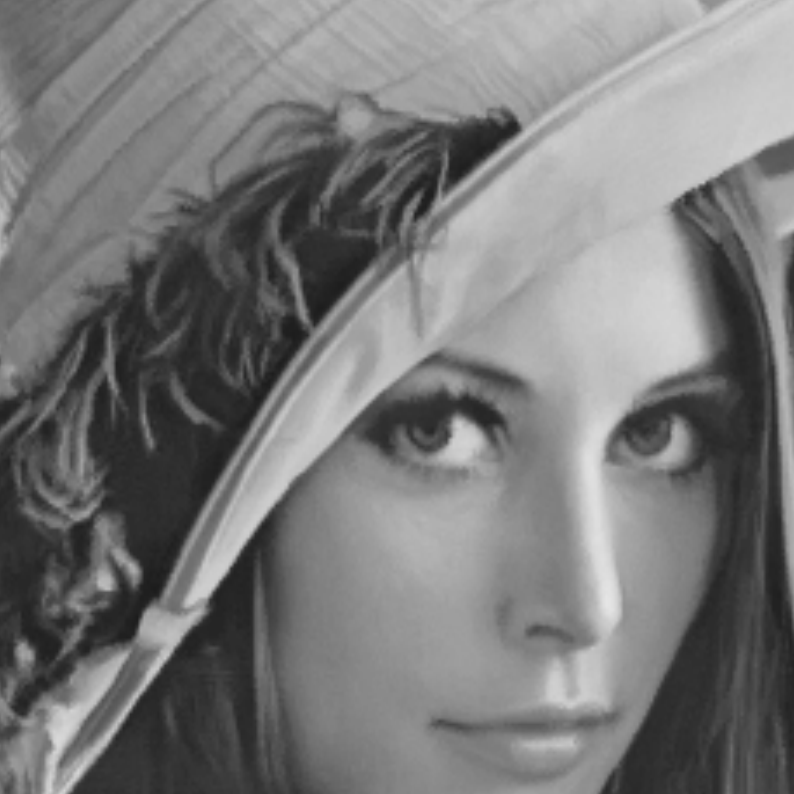}
&\includegraphics[trim = 5cm 5cm 5cm 5cm, clip, width=0.30\linewidth]{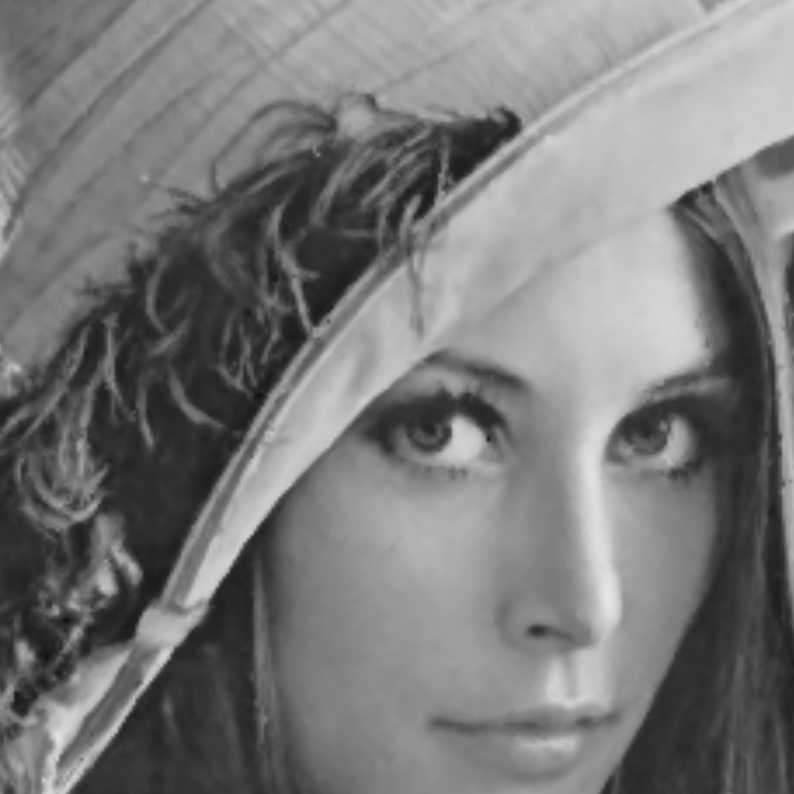}
 \\
 {\small ROLD-EPR PSNR = 34.87 }& {\small PARIGI PSNR = 35.45}& {\small PWMF PSNR = 35.90 }\\
\end{tabular}
\caption{Comparison of the performances of TriF \cite{garnett2005universal}, ROLD-EPR \cite{dong2007detection}, PARIGI \cite{delon2013patch} and our filter PWMF for removing impulse noise with $p=0.2$ for Lena}
\label{figimp}
\end{center}
\end{figure}

\begin{figure}
\begin{center}
\begin{tabular}{ccc}
\includegraphics[trim = 5cm 5cm 5cm 5cm, clip, width=0.30\linewidth]{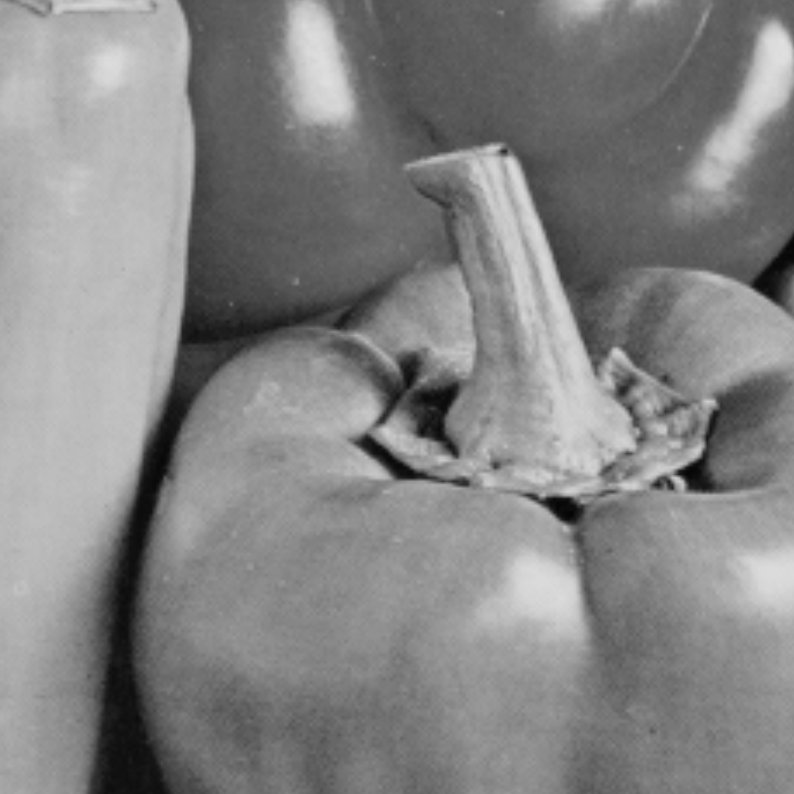}&
\includegraphics[trim = 5cm 5cm 5cm 5cm, clip, width=0.30\linewidth]{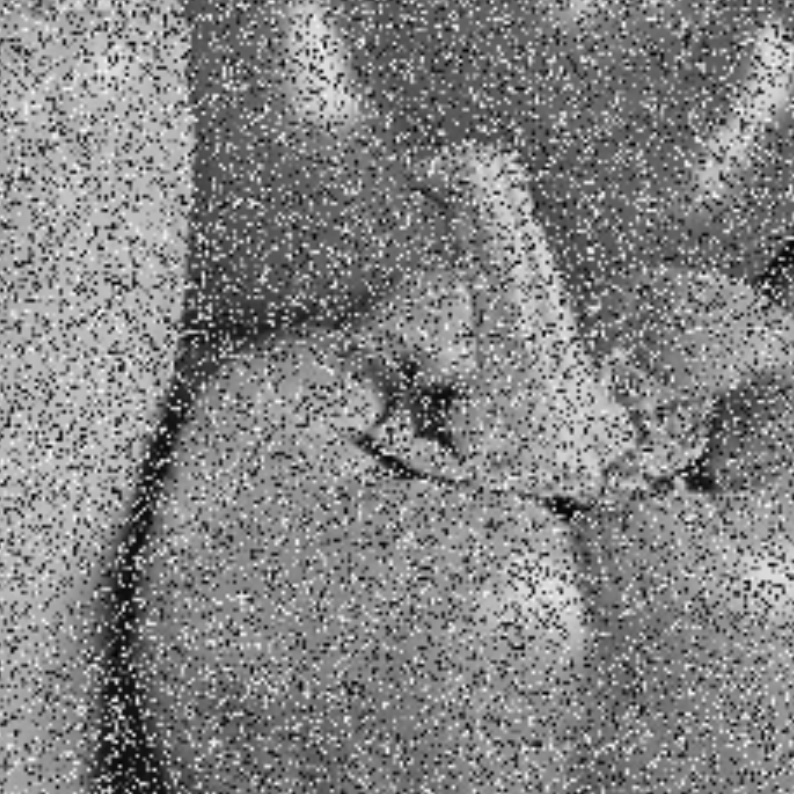} &
\includegraphics[trim = 5cm 5cm 5cm 5cm, clip, width=0.30\linewidth]{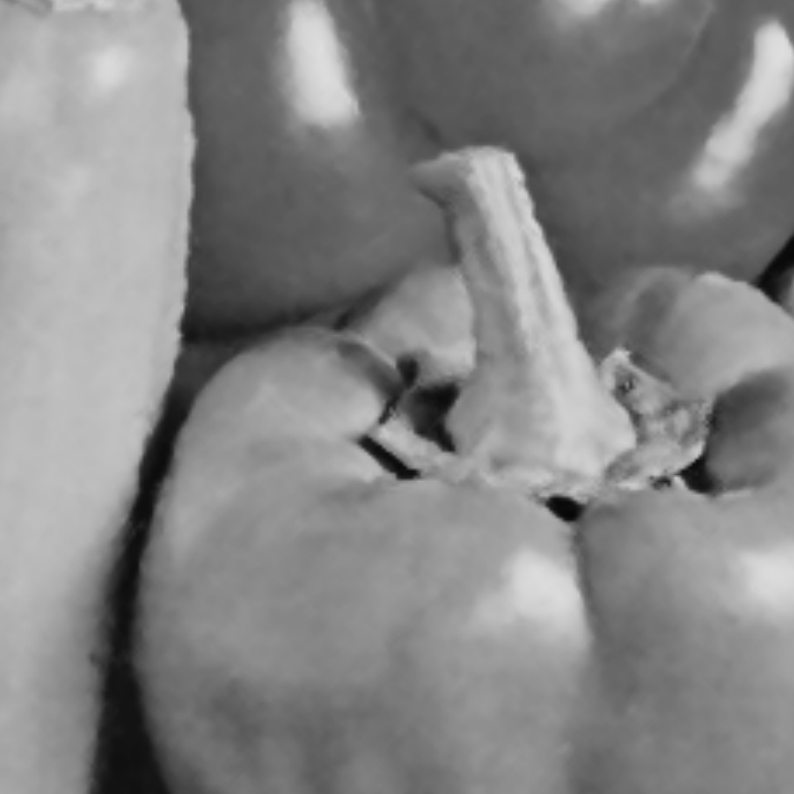}
\\
{\small Original } & {\small Noisy $p$ = 0.4 }&{\small TriF PSNR = 31.27 }\\
\includegraphics[trim = 5cm 5cm 5cm 5cm, clip, width=0.30\linewidth]{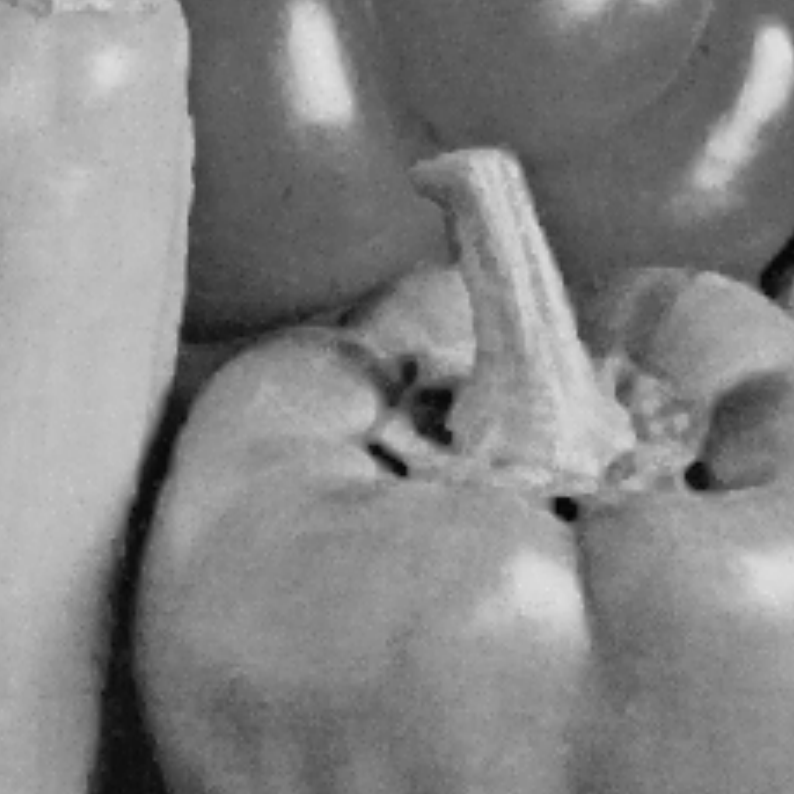} 
&\includegraphics[trim = 5cm 5cm 5cm 5cm, clip, width=0.30\linewidth]{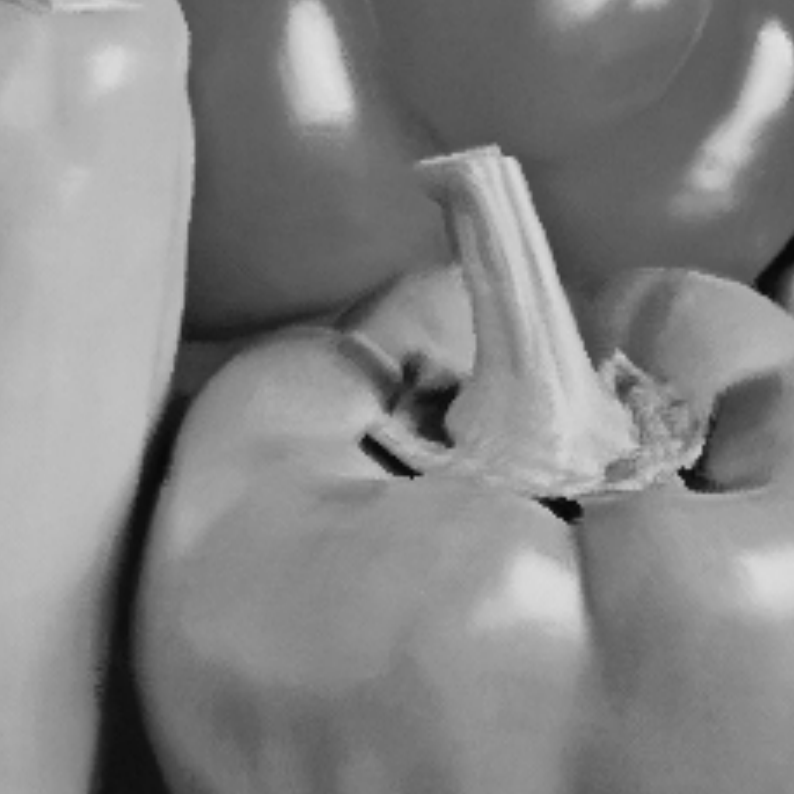}&
\includegraphics[trim = 5cm 5cm 5cm 5cm, clip, width=0.30\linewidth]{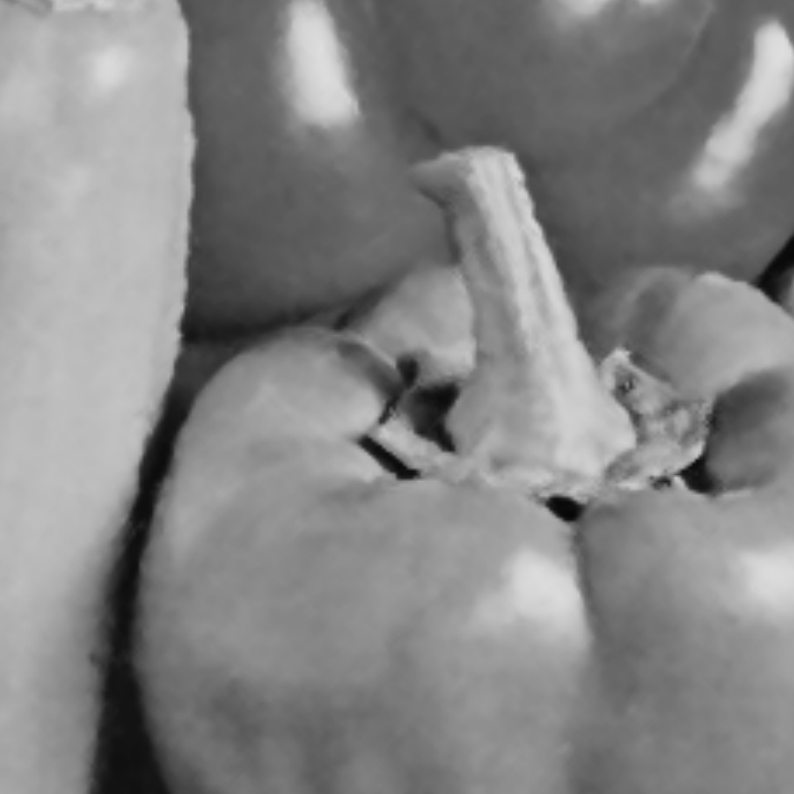}
\\
 {\small ROLD-EPR PSNR = 31.12 }& {\small PARIGI PSNR = 31.63}&{\small PWMF PSNR = 31.95 }\\
\end{tabular}
\caption{Comparison of the performances of TriF \cite{garnett2005universal}, ROLD-EPR \cite{dong2007detection}, PARIGI \cite{delon2013patch} and our filter PWMF for removing impulse noise with $p=0.4$ for Peppers512}
\label{figimp2}
\end{center}
\end{figure}

Table $\ref{psnrimp}$  shows the performances of PWMF for removing impulse noise  by comparing with TriF \cite{garnett2005universal}, ROLD-EPR \cite{dong2007detection}, PARIGI  \cite{delon2013patch}, and NLMixF \cite{huliliu}. 
For ease of comparison, in this and  following tables, we show in bold the best results and the results where the differences from the best ones are less than 0.1dB. We can see that our filter PWMF attains the best performance in term of PSNR. Some visual comparisons are shown in Figs \ref{figimp} and \ref{figimp2}.
% from which we can see that all filter perform rather well. 
Carefully comparing these images, we observe that TriF loses some small textured details, while ROLD-EPR is not smooth enough. PARIGI and PWMF show better results. 

\begin{figure}
\begin{center}
\begin{tabular}{ccc}
 \includegraphics[width=0.30\linewidth]{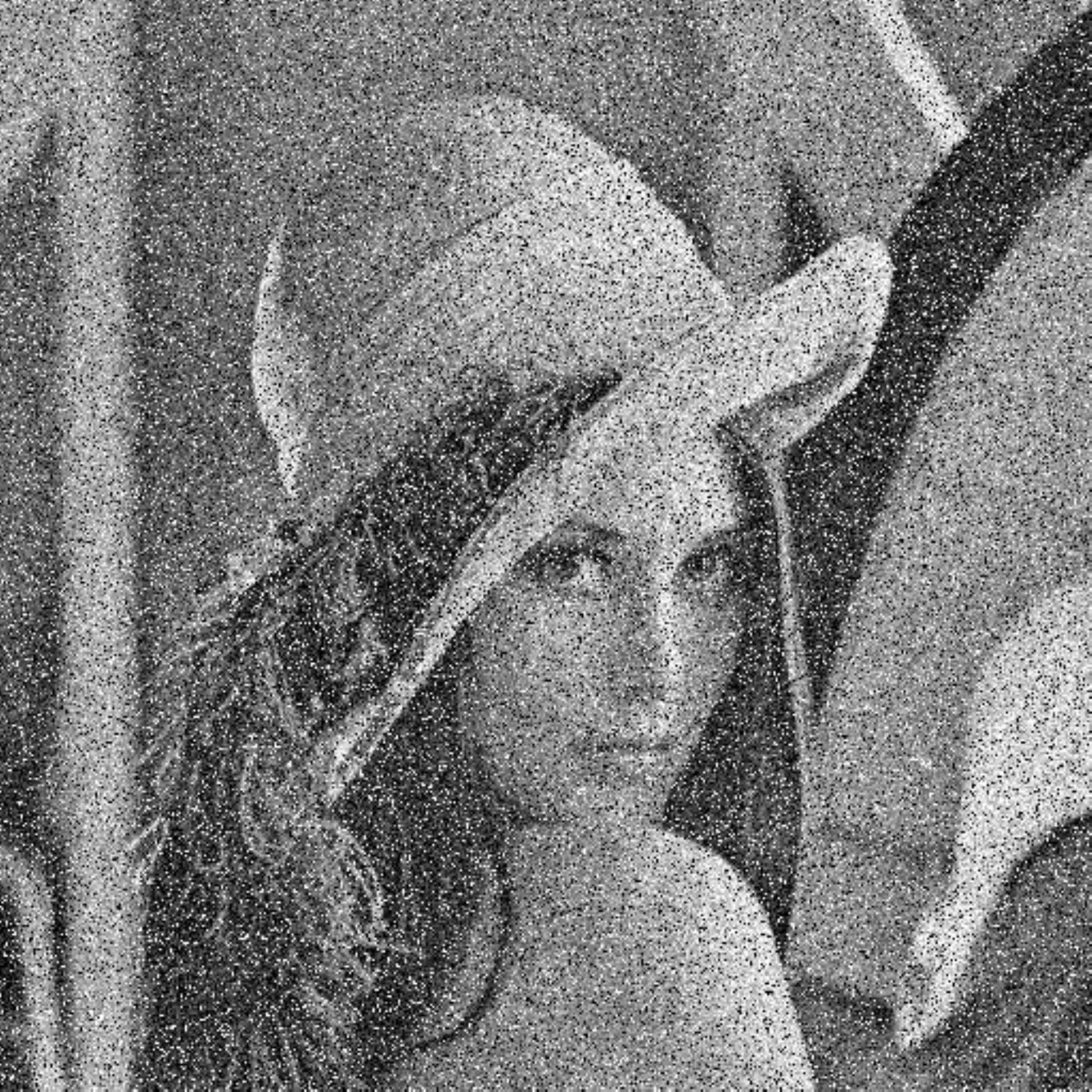}&
\includegraphics[width=0.30\linewidth]{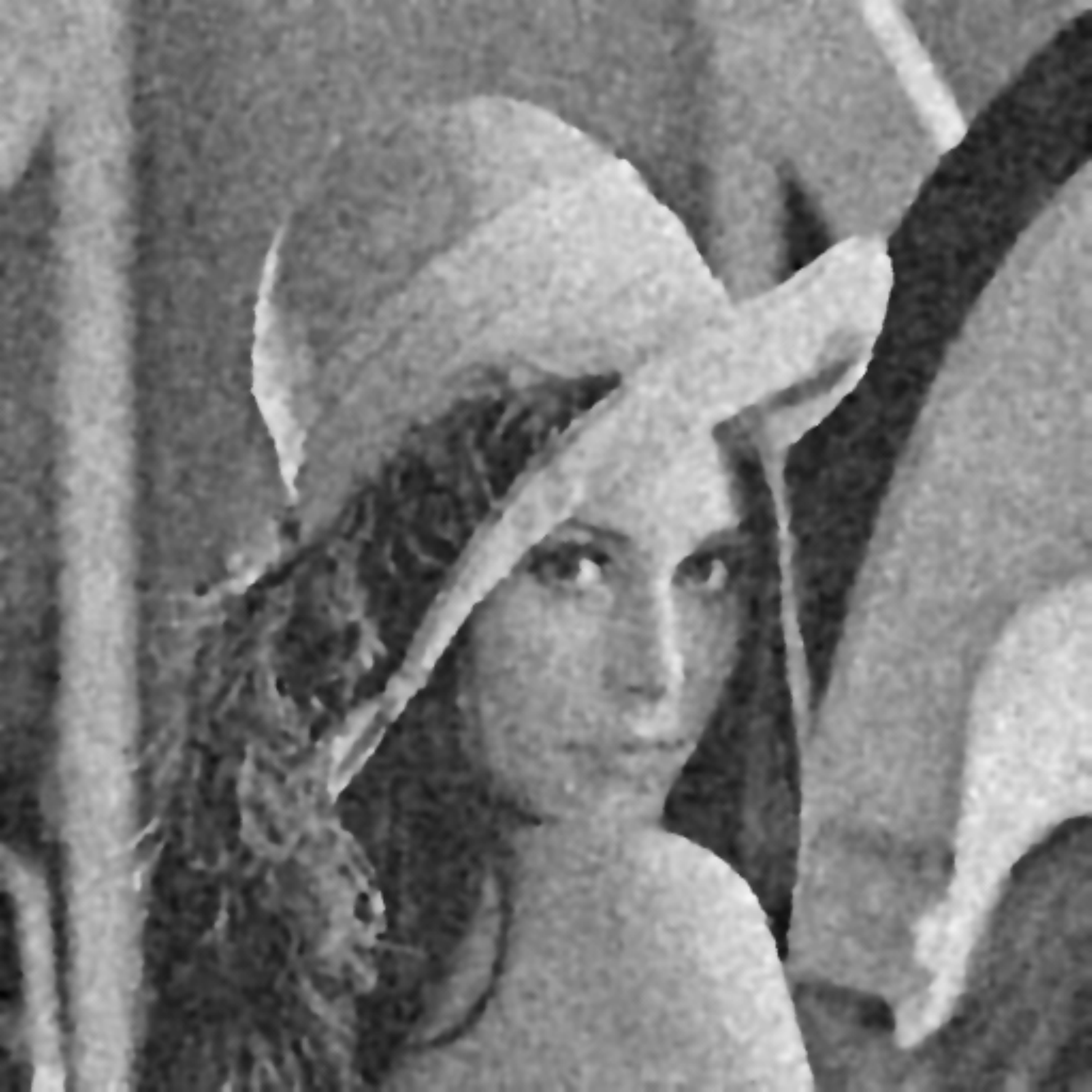}&
\includegraphics[width=0.30\linewidth]{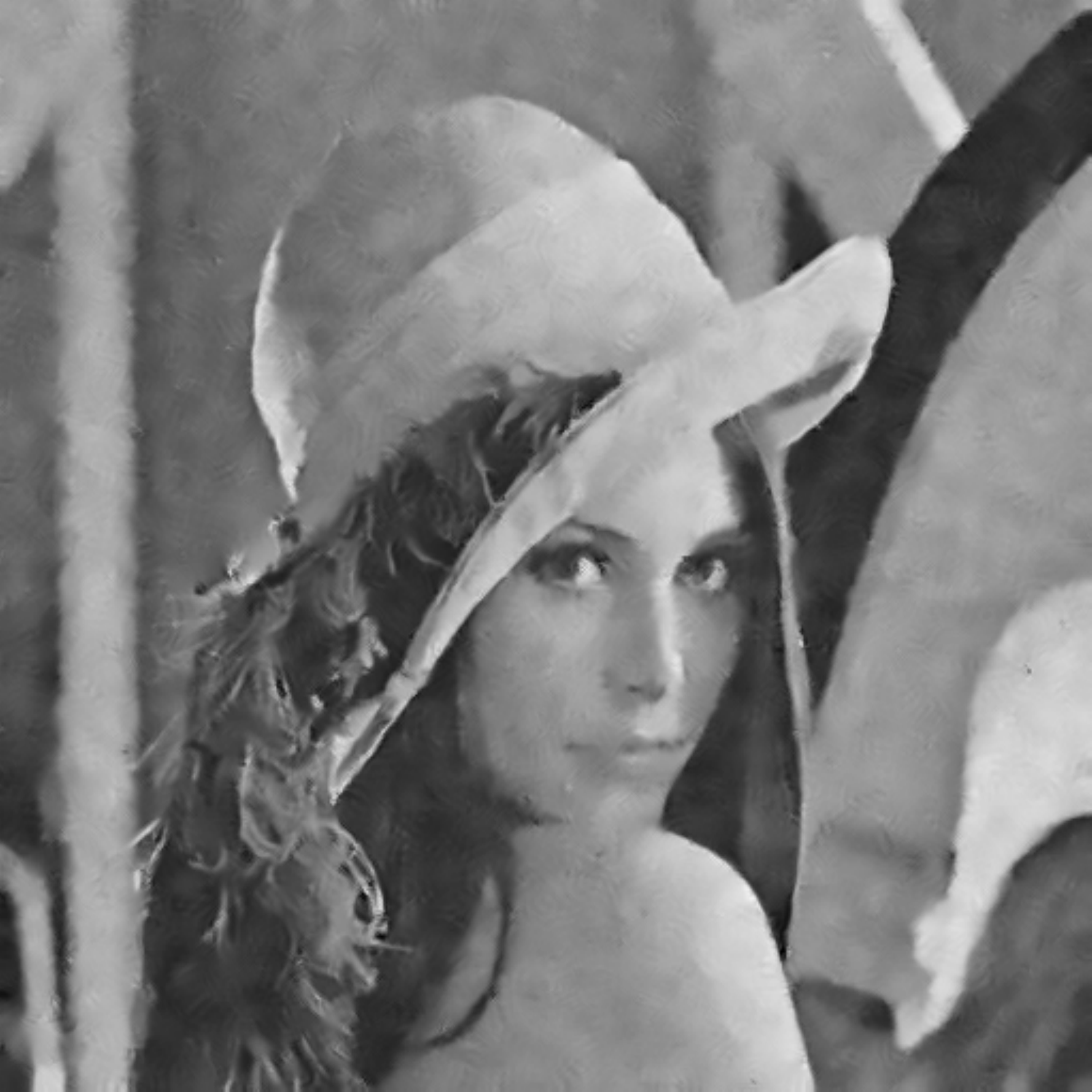}\\
{\small Noisy $\sigma = 30, p = 0.2$} & {\small TriF PSNR = 26.54} &{\small PWMF PSNR = 28.67}\\
\includegraphics[width=0.30\linewidth]{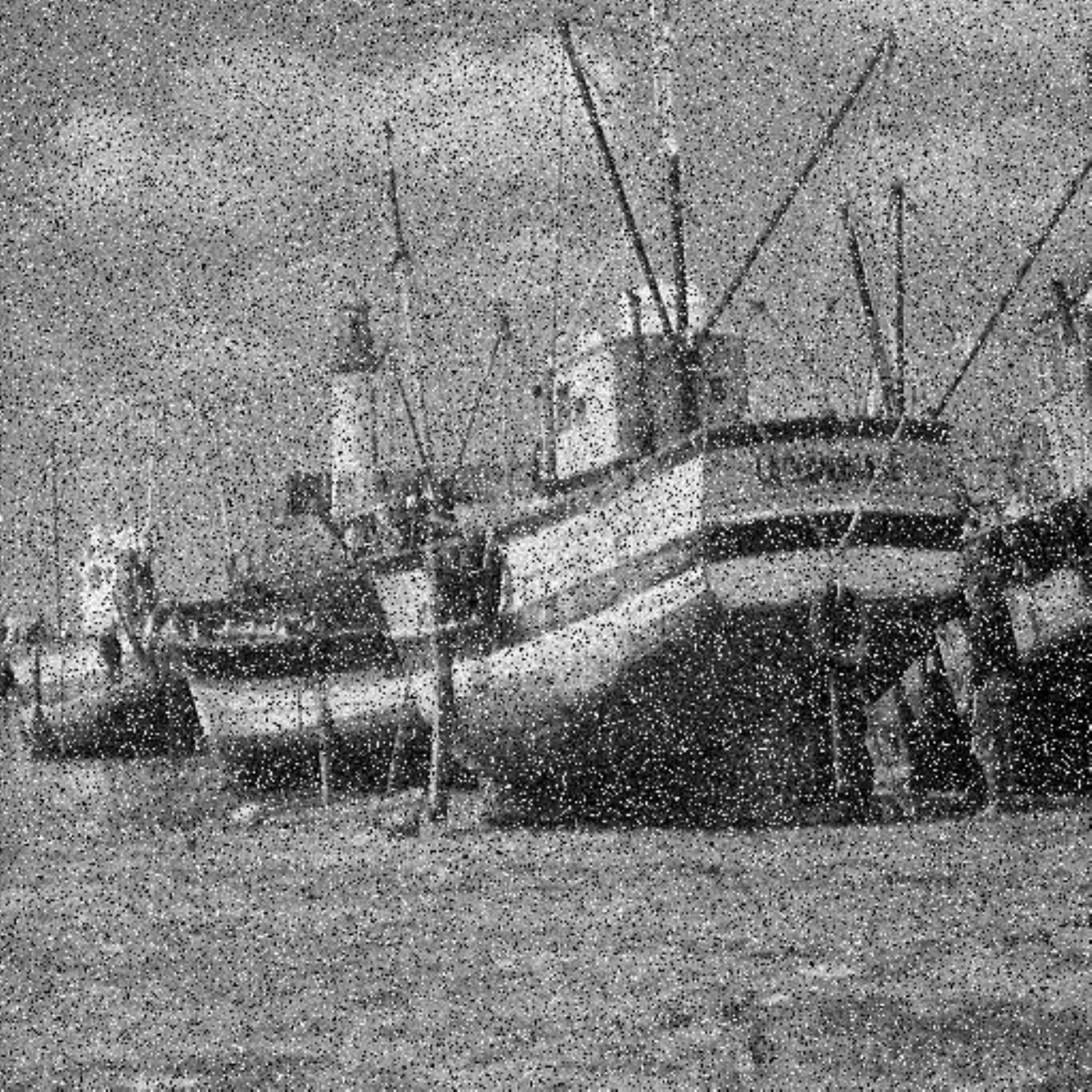}&
\includegraphics[width=0.30\linewidth]{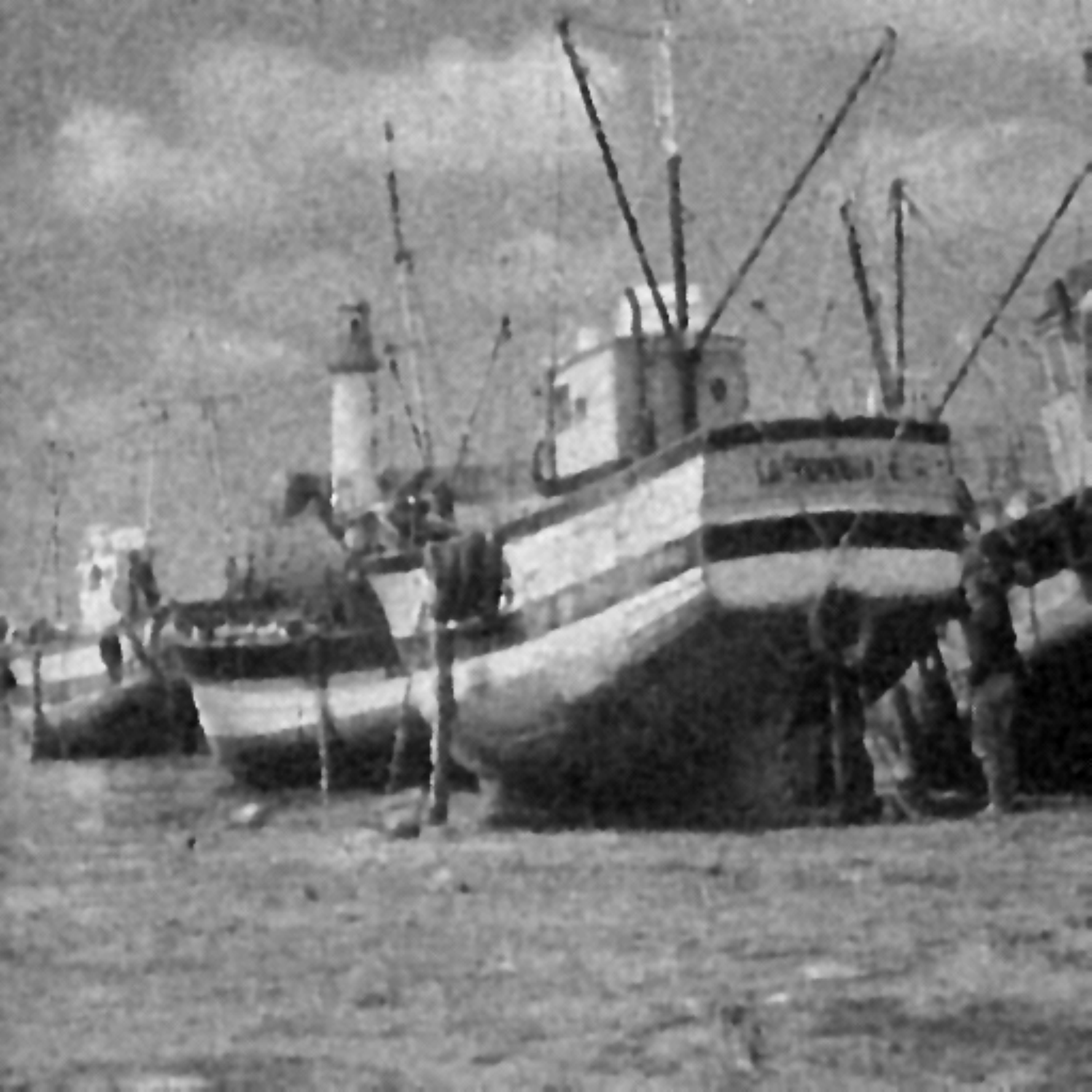}&
\includegraphics[width=0.30\linewidth]{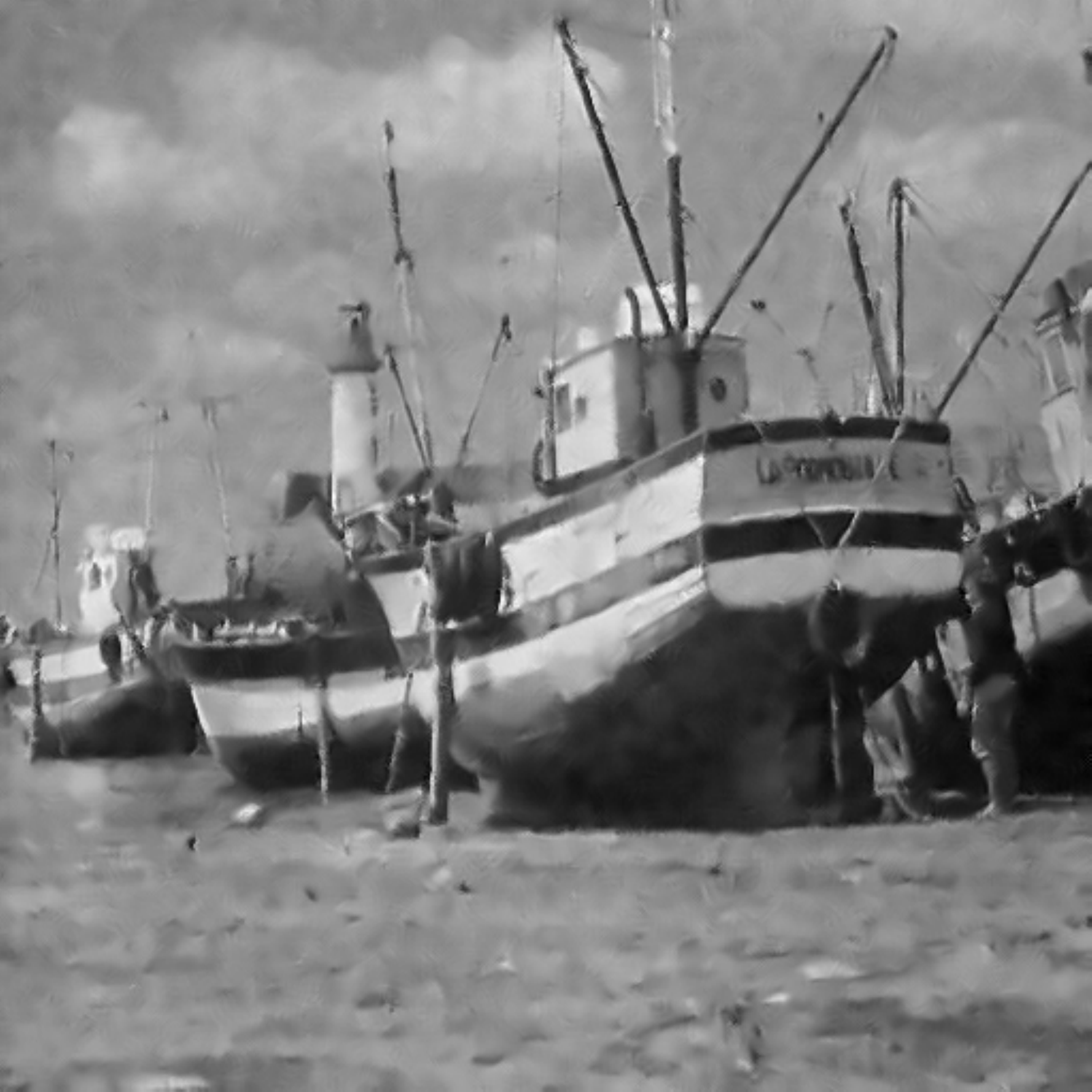}\\
{\small Noisy $\sigma = 20, p = 0.2$ }&{\small TriF PSNR = 26.40}&{\small PWMF PSNR = 27.74}
\end{tabular}
\caption{Comparison of the performances of TriF \cite{garnett2005universal} and our filter PWMF for removing mixed noise}% with $\sigma=30, p=0.2$ and  $\sigma=20, p=0.2$.}

\label{figmix1}
\end{center}
\end{figure}

\begin{figure}
\begin{center}
\begin{tabular}{ccc}
\includegraphics[trim = 5cm 5cm 5cm 5cm, clip, width=0.30\linewidth]{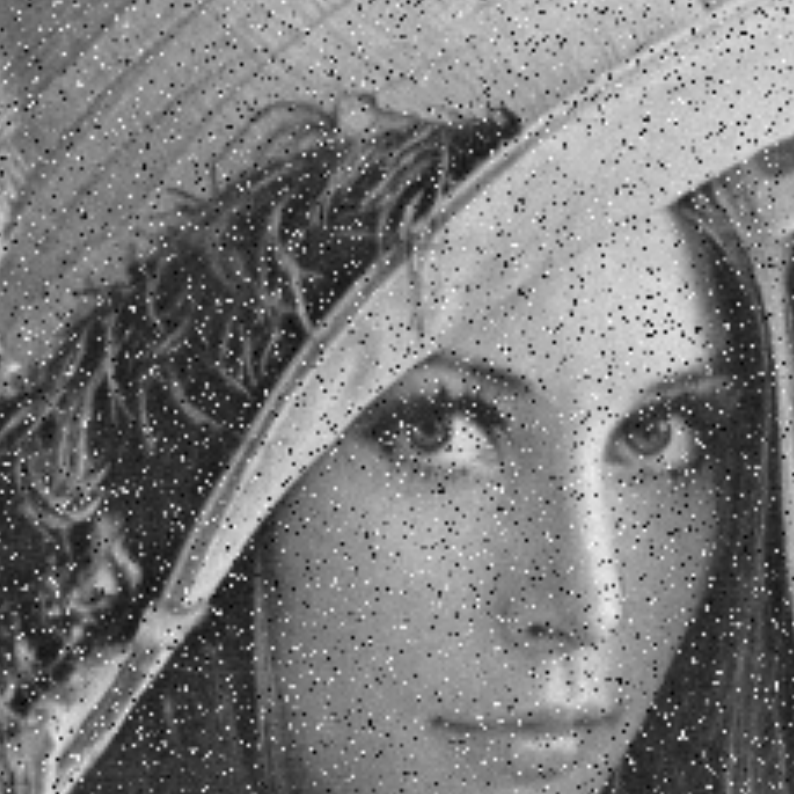}&
\includegraphics[trim = 5cm 5cm 5cm 5cm, clip, width=0.30\linewidth]{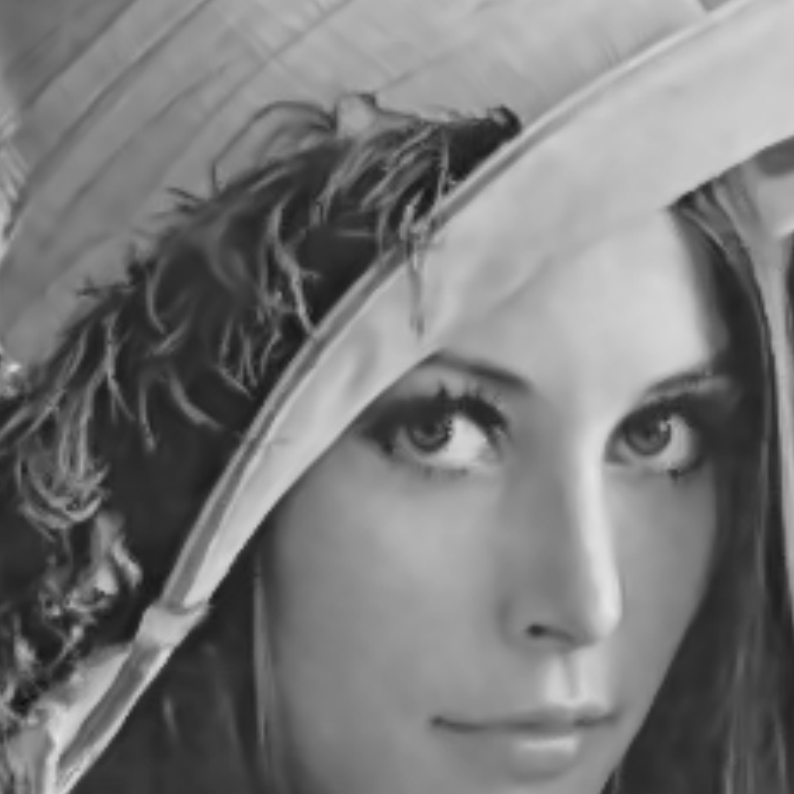}&
\includegraphics[trim = 5cm 5cm 5cm 5cm, clip, width=0.30\linewidth]{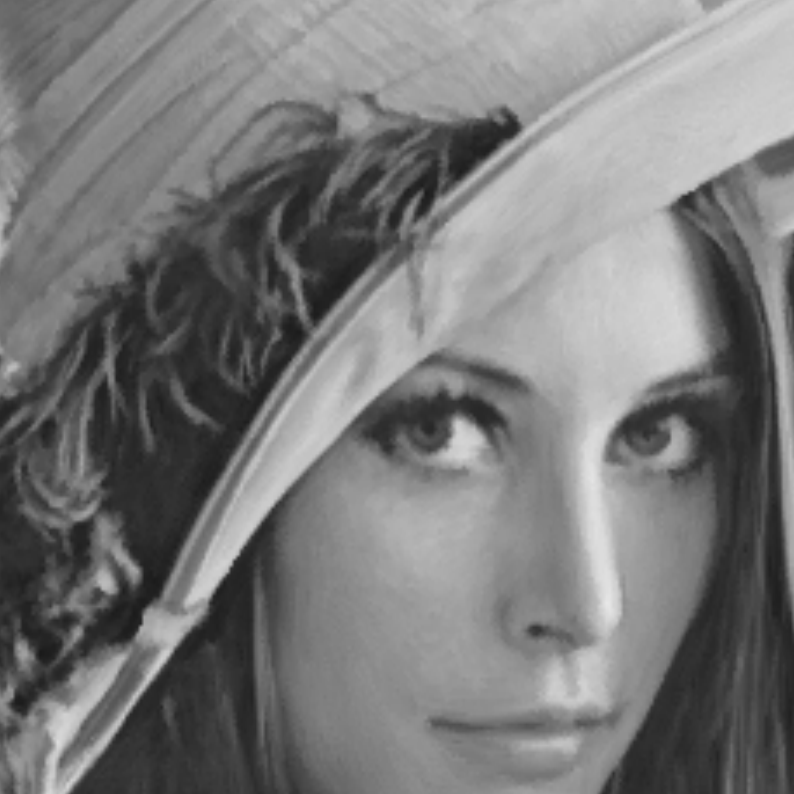}\\
{\small Noisy $p=0.1, \sigma=5$}&{\small PWMF PSNR = 35.80}& {\small PARIGI PSNR = 34.72}\\
\includegraphics[trim = 5cm 5cm 5cm 5cm, clip, width=0.30\linewidth]{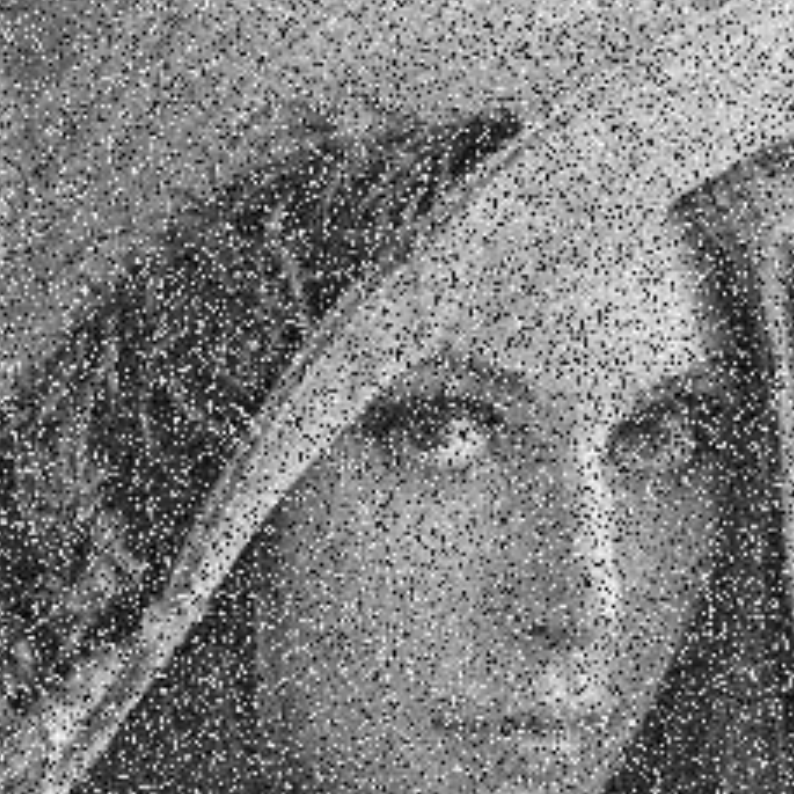}&
\includegraphics[trim = 5cm 5cm 5cm 5cm, clip, width=0.30\linewidth]{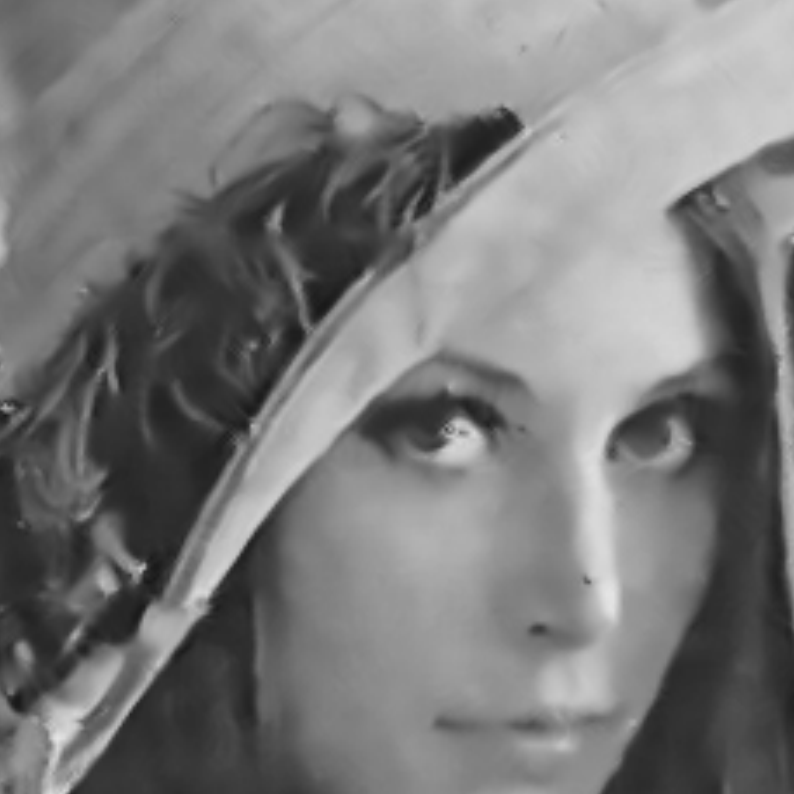}&
\includegraphics[trim = 5cm 5cm 5cm 5cm, clip, width=0.30\linewidth]{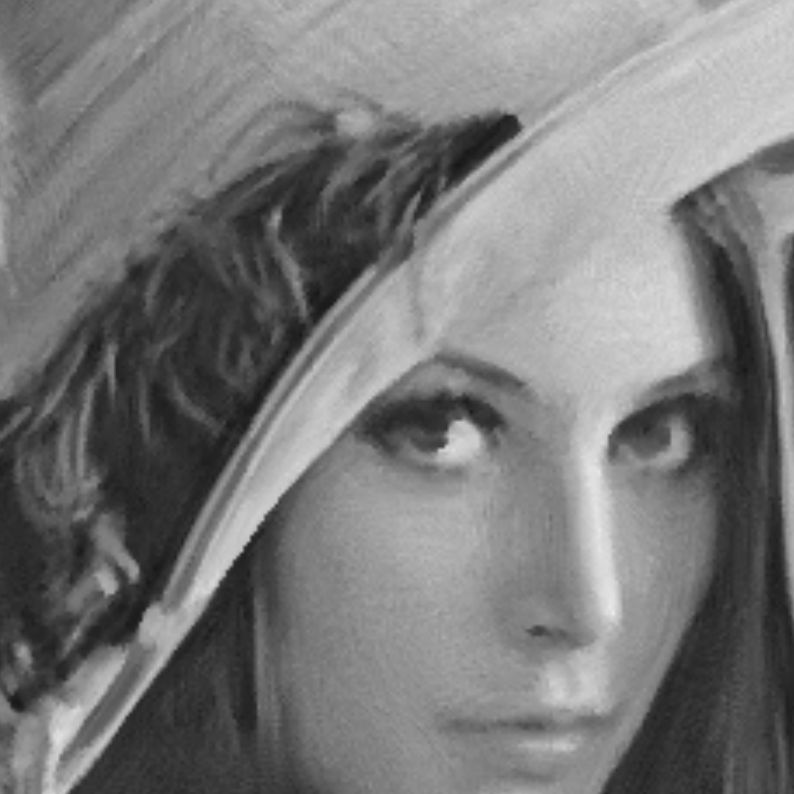}\\
{\small Noisy  $p=0.3, \sigma=15$}&{\small PWMF PSNR = 30.25}& {\small PARIGI PSNR = 29.22}\\
\end{tabular}
\caption{Comparison of the performances of our filter PWMF  and PARIGI \cite{delon2013patch}  for removing   mixed noise with Lena}
\label{figdelon}
\end{center}
\end{figure}

Different papers consider different mixtures of Gaussian noise and impulse noise. We show the performance of PWMF for removing mixed noise in Tables $\ref{psnrmix}$, $\ref{psnrmix_delon}$,  $\ref{tablecom}$, and \ref{tablemnf} by comparing it with TriF \cite{garnett2005universal}, NLMixF \cite{huliliu}, PARIGI \cite{delon2013patch} IPAMF+BM \cite {yang2009mixed}, Xiao \cite{xiao2011restoration}, MNF \cite{li2011new}, and Zhou \cite{zhou2013restoration}. 
 All these comparisons show good performance of our filter except for Barbara when comparing with PARIGI. Our method does not work very well as PARIGI for Barbara, because the ROAD statistics is a very local statistics and  can not use the redundancy of this image very well to  detect impulse noisy pixels, while PARIGI is particularly powerful for the restoration of textured regions. From  Fig. $ \ref{figmix1}$, we can see that the results of our filter are visually better than TriF.   From Fig. $ \ref{figdelon}$, we can see that  when the standard deviation $\sigma$ is high, our filter is smoother than PARIGI, while PARIGI seems to preserve more weak textured  details, but  it has evident artifacts throughout the whole image (see the electronic version of this paper at full resolution).

\begin{table}
 \begin{center}
 \caption{ PSNR values (dB)
to remove impulse noise for TriF \cite{garnett2005universal}, ROLD-EPR \cite{dong2007detection}, PARIGI \cite{delon2013patch}, NLMixF \cite{huliliu} and our filter PWMF}
 {\footnotesize
%%-------------------------from tabletex
\begin{tabular}{ccccc}
\hline\noalign{\smallskip}
 Lena  & $ p=0.2 $ & $ p= 0.3 $ & $ p=  0.4$ & $ p=0.5$\\ 
\noalign{\smallskip}
\hline
\noalign{\smallskip}
TriF &34.75  &  32.54 &   31.28   & 29.37 \\
ROLD-EPR &34.87    &   32.08   &     30.81   &     29.51\\
PARIGI & 35.45 & -& 31.75 &- \\
NLMixF &35.69   & 33.13  &  31.69  &  29.87 \\
PWMF& \textbf{35.90} &\textbf{33.45} &\textbf{31.98} &\textbf{30.17} \\
\end{tabular}

\vskip3mm
\begin{tabular}{ccccc}
\hline\noalign{\smallskip}
 Bridge  & $ p=0.2 $ & $ p= 0.3 $ & $ p=  0.4$ & $ p=0.5$\\ 
\noalign{\smallskip}
\hline\noalign{\smallskip}
TriF & 26.81  &  25.25&    24.41  &  23.35 \\
ROLD-EPR &27.60     & 25.58    &   24.42    &    23.45 \\
PARIGI & 27.68 &- & \textbf{24.80} & - \\
NLMixF &27.77    &25.54   & 24.45  &  23.33\\
PWMF&\textbf{28.10} &\textbf{26.11} &\textbf{24.74}& \textbf{23.64} \\
\end{tabular}

\vskip3mm
\begin{tabular}{ccccc}
\hline\noalign{\smallskip}
 Peppers256  & $ p=0.2 $ & $ p= 0.3 $ & $ p=  0.4$ & $ p=0.5$\\ 
\noalign{\smallskip}
 \hline\noalign{\smallskip}
 TriF & 30.55 &   28.81 &   27.67 &   25.98\\
 ROLD-EPR &31.03   &    28.10     &  27.34   &     25.96\\
 PARIGI & - &- & - & - \\
 NLMixF &31.78 &   29.38   & 28.10  &  26.47 \\
 PWMF&\textbf{31.93}   & \textbf{29.64} &    \textbf{28.34} &    \textbf{26.78}

 \end{tabular}

\vskip3mm
\begin{tabular}{ccccc}
\hline\noalign{\smallskip}
 Peppers512  & $ p=0.2 $ & $ p= 0.3 $ & $ p=  0.4$ & $ p=0.5$\\ 
\noalign{\smallskip}
\hline\noalign{\smallskip}
TriF & 34.52 & 31.93& 31.27 & 29.76\\
ROLD-EPR & 34.46 & 32.31 &31.12 &30.03 \\
PARIGI & 34.75 &-& 31.63 & - \\
NLMixF &34.77 & \textbf{32.56} &31.73&30.23 \\
PWMF&\textbf{35.08}& \textbf{32.59}&\textbf{31.95}&\textbf{30.38}

\end{tabular}

\vskip3mm
\begin{tabular}{ccccc}
\hline\noalign{\smallskip}
 Boats  & $ p=0.2 $ & $ p= 0.3 $ & $ p=  0.4$ & $ p=0.5$\\ 
\noalign{\smallskip}
\hline\noalign{\smallskip}
TriF & 30.22 &   28.55 &   27.52&    26.10\\
ROLD-EPR &30.75   &    28.19    &    26.95  &      25.91\\
PARIGI & 31.21 &-& 27.56 & - \\
NLMixF &31.32  &  29.01  &  27.42  &  26.10 \\
PWMF&\textbf{31.83} &\textbf{29.58} &\textbf{27.67} &\textbf{26.48}\\
\end{tabular}

\vskip3mm
%%-------------------------end tabletex

 }

\label{psnrimp}
\end{center}
\end{table}

%fig2

%fig3

\begin{table}
\begin{center}
\caption{ PSNR values (dB)
to remove mixed noise for TriF \cite{garnett2005universal}, NLMixF \cite{huliliu} and our filter PWMF. P256 and P512 represent Peppers256 and Peppers512 respectively. For each PSNR value, the corresponding  level of Gaussian noise and impulse noise are shown in the top and left of the table }% F2: NLMixF-2, F3: NLMixF-3}
\addtolength{\tabcolsep}{-2pt}
 {\footnotesize
%%-------------------------from tabletex
\begin{tabular}{c|ccc|ccc|ccc}
\hline
 $p$ & \multicolumn{3}{c|}{$\sigma=10$}        &  \multicolumn{3}{c|}{$\sigma=20$} &      
\multicolumn{3}{c}{$\sigma=30$}  \\
\hline
Lena & TriF    &   NLMixF      & PWMF  & TriF   &   NLMixF       & PWMF  &
TriF   &   NLMixF   & PWMF\\
$0.2$&31.70  & \textbf{32.85} &\textbf{32.93}  &28.75 & \textbf{30.51} &\textbf{30.47}   &26.54 &\textbf{28.70} &\textbf{28.67}   \\
$0.3$&30.77 & \textbf{31.28} &\textbf{31.30} &28.01 & \textbf{29.41} &\textbf{29.38 }   &25.82& \textbf{27.68} &\textbf{27.65}   \\
\hline
Bridge & TriF  &   NLMixF        & PWMF & TriF  &   NLMixF        & PWMF&  
TriF     &   NLMixF     & PWMF\\
$0.2$&25.28 & 26.14 &\textbf{26.35}  &23.84 &\textbf{24.48} & \textbf{24.53}   &22.50 & \textbf{23.33} & \textbf{23.35}   \\
$0.3$&24.66 & 24.73& \textbf{25.00} & 23.34 & 23.59 &\textbf{23.70}     &22.08 &\textbf{22.66}&\textbf{22.72}  \\
\hline
P256 & TriF    &   NLMixF      & PWMF&  TriF    &   NLMixF      & PWMF&  
TriF   &   NLMixF       & PWMF \\
$0.2$ &29.09 &\textbf{30.57} &  \textbf{30.65}   & 26.82  & \textbf{28.56} &\textbf{28.53}    & 24.70 & \textbf{26.73} &\textbf{26.71}  \\
$0.3$& 27.89 & \textbf{28.68} &\textbf{28.73}   & 26.01   & \textbf{27.16} &\textbf{27.13}  & 23.97 & \textbf{25.55} &\textbf{25.57}     \\
\hline
P512 & TriF    &   NLMixF      & PWMF&  TriF    &   NLMixF      & PWMF&  
TriF   &   NLMixF       & PWMF \\
$0.2$&31.84  & \textbf{32.64}  &  \textbf{32.70} & 29.01 &  \textbf{30.74} &  \textbf{30.69} & 26.82 &  \textbf{29.04} &\textbf{29.07} \\
$0.3$&30.92  &  \textbf{31.06} & 30.93 & 28.27 & \textbf{29.55} & 29.39 & 26.01 & \textbf{27.92} & \textbf{27.86} \\
\hline 
Boats & TriF    &   NLMixF     & PWMF&   TriF  &   NLMixF        & PWMF& 
TriF    &   NLMixF      & PWMF \\
$0.2$& 28.37 & 29.75 &\textbf{29.91}   & 26.40 & \textbf{27.71}  & \textbf{27.74}   &24.58 & \textbf{26.23} &\textbf{26.23}  \\
$0.3$&27.63 & 28.15 &\textbf{28.38}    &25.84 & \textbf{26.66} &\textbf{26.75 }    & 24.09 & \textbf{25.44} &\textbf{25.48} 

\end{tabular}

\vskip3mm
%%-------------------------end tabletex

 }

\label{psnrmix}
\end{center}
\end{table}

\begin{table}
 \begin{center}
 \caption{ PSNR values (dB)
to remove mixed noise for PARIGI \cite{delon2013patch} and our filter PWMF}
 {\footnotesize
%%-------------------------from tabletex
\begin{tabular}{ccccc}
\hline\noalign{\smallskip}
 $p=0.1\: \sigma=5$  & Lena & Barbara & Cameraman & Boat\\ 
\noalign{\smallskip}
\hline\noalign{\smallskip}
PARIGI &34.72    &   \textbf{31.55}   &     34.98  &     31.41\\%\noalign{\smallskip}
PWMF& \textbf{35.80} &{30.86} &\textbf{35.80} &\textbf{32.60} \\
\end{tabular}

\vskip3mm
\begin{tabular}{ccccc}
\hline\noalign{\smallskip}
$p=0.3 \:\sigma=15$  & Lena & Barbara & Cameraman & Boat \\ \noalign{\smallskip}
\hline\noalign{\smallskip}
PARIGI &29.22     & \textbf{27.33}    &   28.59    &   26.57 \\
PWMF&\textbf{30.25} &25.58 &\textbf{30.27}& \textbf{27.45} \\
\end{tabular}

%%-------------------------end tabletex

 }

\label{psnrmix_delon}
\end{center}
\end{table}

\begin{table}
\begin{center}
\caption{ PSNR values (dB) for mixed noise removal  with  (Xiao) \cite{xiao2011restoration},   (IPAMF+BM) \cite{yang2009mixed}, (Zhou) \cite{zhou2013restoration} and our filter PWMF}
 \vskip3mm
 {\footnotesize
%%-------------------------from tabletex
\begin{tabular}{cccc}
\hline\noalign{\smallskip}
 Lena $\sigma=10$ & $ p=0.1 $  &$ p= 0.2 $ & $p=0.3$ \\ 
\noalign{\smallskip}
\hline \noalign{\smallskip}
 Xiao & 32.75 & 31.66 & 30.42 \\
  IPAMF+BM & 33.61 & 32.12 & 30.69 \\
Zhou & $\mathbf{34.25}$ & 32.68 & $\mathbf{31.21}$ \\
PWMF & ${34.10}$ & $\mathbf{32.93}$ & $\mathbf{31.30}$ \\
%PWMF & $\mathbf{34.10}$ & $\mathbf{32.78}$ & $\mathbf{31.47}$ \\
\end{tabular}
}
\label{tablecom}
\end{center}
\end{table}

\begin{table}
\begin{center}
 \caption{ PSNR values (dB) for mixed noise removal with  MNF \cite{li2011new}  and our filter PWMF}
 {\footnotesize
%%-------------------------from tabletex
\begin{tabular}{cccc}
\hline\noalign{\smallskip}
 Lena & $\sigma=10, p=0.2$ & $\sigma=20, p=0.2 $  &$\sigma=30, p= 0.2 $\\ % & $\sigma=10, p=0.4$ \\ 
\noalign{\smallskip}
\hline \noalign{\smallskip}
 %MNF & 31.96 & 29.64 & & 30.42 \\
 MNF & 31.63 & 29.33 &28.40 \\ 
%PWMF & $\mathbf{32.85}$ & $\mathbf{30.51}$ &$\mathbf{28.70}$ &  $\mathbf{29.94}$ \\$\mathbf{29.94}$ & 28.80
PWMF & $\mathbf{32.93}$ & $\mathbf{30.47}$ &$\mathbf{28.67}$ \\  
\end{tabular}
}
\label{tablemnf}
\end{center}
\end{table}

%fig4

%fig5

Finally, we compare the CPU time of  TriF \cite{garnett2005universal}, NLMixF \cite{huliliu}, and our method PWMF for removing mixed nose in seconds in the platform of MATLAB R2011a with unoptimized mex files. The computer is equipped with 2.13GHZ Intel (R) Core (TM) i3 CPU and 3.0 GB memory. The results are presented in Table \ref{time}, which demonstrate that PWMF is rather fast: much faster than NLMixF and even faster than TriF when the noise level is low, thanks to the simplified   joint impulse factor $F\big(k, \mathcal{T}(k)\big)$ defined in (\ref{sigsm}). 
The results also show that our method is faster than Zhou \cite{zhou2013restoration}.

\begin{table}
 \begin{center}
 \vskip3mm
\caption{Time(s) for TriF  \cite{garnett2005universal}, NLMixF \cite{huliliu}, and PWMF}
 {\footnotesize
%%-------------------------from tabletex
\begin{tabular}{ccccc}
\hline\noalign{\smallskip}
Image & Noise levels & TriF & NLMixF & PWMF\\
\noalign{\smallskip}
\hline\noalign{\smallskip}
 Lena  & $ \sigma=10, p=0.2  $ & 6.9  &  60.6 & 5.4 \\ 
%\hline
Lena  & $ \sigma=20, p=0.3  $ & 7.9  & 170.0 & 16.7 \\ 
\end{tabular}
}
\label{time}
\end{center}
\end{table}

\section{Conclusions and Further Work}
Two convergence theorems,  one for the almost sure convergence and the other for the convergence in law, are established    to show  the rate of convergence of NL-means \cite{buades2005review}.
The notion of degree of similarity is also introduced to describe the influence of the proportion  of similar patches in  the application of NL-means. Based on the convergence theorems,  a new filter called patch-based weighted means filter (PWMF) is proposed to remove mixed noise, leading to an extension  of NL-means.  The choice of parameters has been carefully discussed. Simulation results show that the new proposed filter is competitive compared to recently developed known algorithms.

As the detection of impulse noise is crucial for removing the noise, which is done by the statistics ROAD \cite{garnett2005universal} in this paper, we could further improve our results by improving ROAD. In the future, we will consider  a semi-local statistics to make use of redundancies of images, which has the possibility to well recover the textured regions. 

%\begin{acknowledgements}
%\end{acknowledgements}

%\begin{acknowledgements}
%The authors would like to thank Raymond H. Chan and Yiqiu Dong  for kindly providing us the ROLD-EPR denoising code.
%\end{acknowledgements}


%\begin{thebibliography}{}
% etc
%\end{thebibliography}

%\bibliographystyle{spmpsci}  
%\bibliography{these}

\end{document}